\documentclass{cta-author}
\usepackage{algorithm}
\usepackage{algorithm}
\usepackage{algorithmic}
\usepackage{subfigure}
\usepackage[colorlinks,linkcolor=black,citecolor=black]{hyperref}
\usepackage{float}
\usepackage{natbib}
\usepackage{multirow}
\usepackage{multicol}
\usepackage{soul} % 导入 soul 包
\usepackage{color, xcolor} % 颜色包，color 必须导入，xcolor 建议导入
{}
{}
{}
\soulregister{\cite}7 % 注册\cite命令
\soulregister{\citep}7 % 注册\citep命令
\soulregister{\citet}7 % 注册\citet命令
\soulregister{\ref}7 % 注册\ref命令
\soulregister{\pageref}7 % 注册\pageref命令
\begin{document}

\title{Clustering-Induced Generative Incomplete Image-Text Clustering(CIGIT-C)}

\author{\au{Xiaoming Su$^{1,2\corr}$}, \au{Dongjin Guo$^{1,2}$},\au{Jiatai Wang$^{1,2}$,}
\au{Limin Liu$^{1,2}$}, }

\address{\add{1}{College of Data Science and Application,Inner Mongolia University of Technology, Huhhot, China}
\add{2}{Big Data Laboratory of Inner Mongolia Discipline Supervision and Investigation, Huhhot, China}
\corr Correspondence
\\
Xiaoming Su, College of Data Science and Application,
Inner Mongolia University of Technology, Huhhot,
China
\\
Email: {xiaoming${\_}$su@foxmail.com}}

\begin{abstract}
The target of image-text clustering (ITC) is to find correct clusters by integrating complementary and consistent information of multi-modalities for these heterogeneous samples. However, the majority of current studies analyse ITC on the ideal premise that the samples in every modality are complete. This presumption, however, is not always valid in real-world situations. The missing data issue degenerates the image-text feature learning performance and will finally affect the generalization abilities in ITC tasks. Although a series of methods have been proposed to address this incomplete image text clustering issue (IITC), the following problems still exist: 1) most existing methods hardly consider the distinct gap between heterogeneous feature domains. 2) For missing data, the representations generated by existing methods are rarely guaranteed to suit clustering tasks. 3) Existing methods do not tap into the latent connections both inter and intra modalities.  In this paper, we propose a Clustering-Induced Generative Incomplete Image-Text Clustering(CIGIT-C) network to address the challenges above. More specifically, we first use modality-specific encoders to map original features to more distinctive subspaces. The latent connections between intra and inter-modalities are thoroughly explored by using the adversarial generating network to produce one modality conditional on the other modality. Finally, we update the corresponding modality-specific encoders using two KL divergence losses. Experiment results on public image-text datasets demonstrated that the suggested method outperforms and is more effective in the IITC job.
\end{abstract}

\maketitle

\section{INTRODUCTION}\label{introduction}

 The two primary ways of displaying information in daily life are visual media and natural language. Therefore, as a fundamental research topic in machine learning and data science communities, ITC has drawn an increasing amount of interest in recent years. It is advantageous for a variety of applications, including vehicle detection \cite{gomaa2020efficient} \cite{gomaa2018real} \cite{gomaa2019robust}, clinical medicine \cite{horne2020challenges}, marketing research, crowdsourced mobile testing \cite{li2021hybrid}, and recommender system \cite{do2020image}. 
 
  Existing ITC methods are divided into two categories in the literature: conventional and deep approaches. The conventional methods, while they have progressed, still rely on manually created features and linear embedding functions. They are unable to grasp the non-linear structure of complex cross-modal data as a consequence.
In DNN-based ITC methods, they are usually classified into the following two categories. The first category employs a two-stage procedure, in which features based on DNN are initially extracted, and then the final clustering results are learned using the conventional clustering method. The second category establishes a clustering loss to steer the multi-modal feature learning process. The success of ITC relies on the assumption of information completeness: instances in all modalities correspond to each other.

\begin{figure}[h]
\centering
\includegraphics[width=0.5\textwidth]{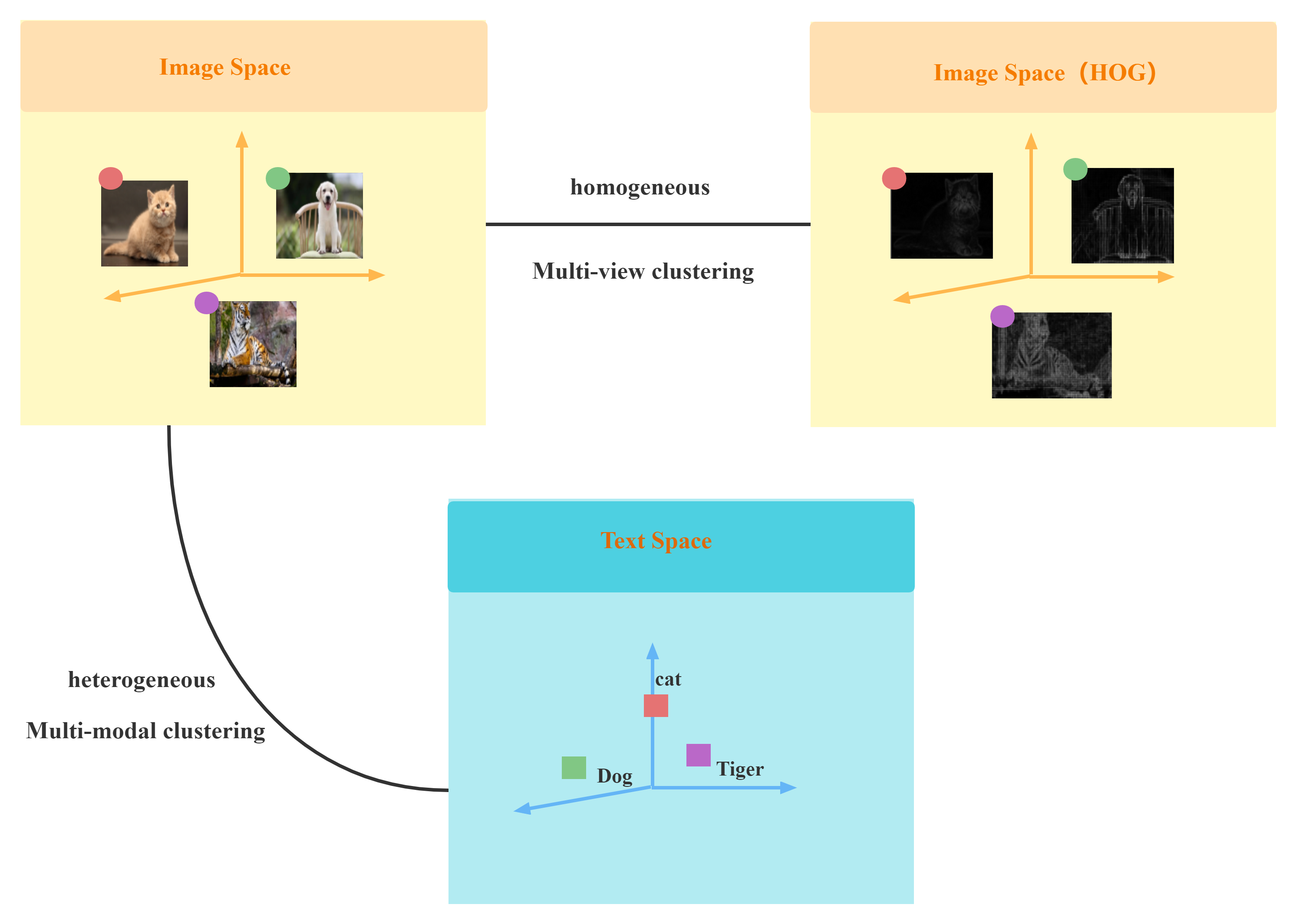}
\caption{The reason for incomplete multi-view clustering approach does not solve the IITC task well. Because the incomplete multi-view approach targets homogeneous modalities, as shown in the two yellow boxes, i.e., (Image Space) and (Image Space (HOG)). IITC essentially solves the problem of heterogeneous modalities, which belong to the paradigm of multi-modal clustering methods, as shown in yellow and blue rectangles in the figure, i.e., (Image Space) and (Text Space).}
\label{Fig1}
\end{figure}

\begin{figure*}[t]
  \begin{center}
  \includegraphics[width=0.90\textwidth]{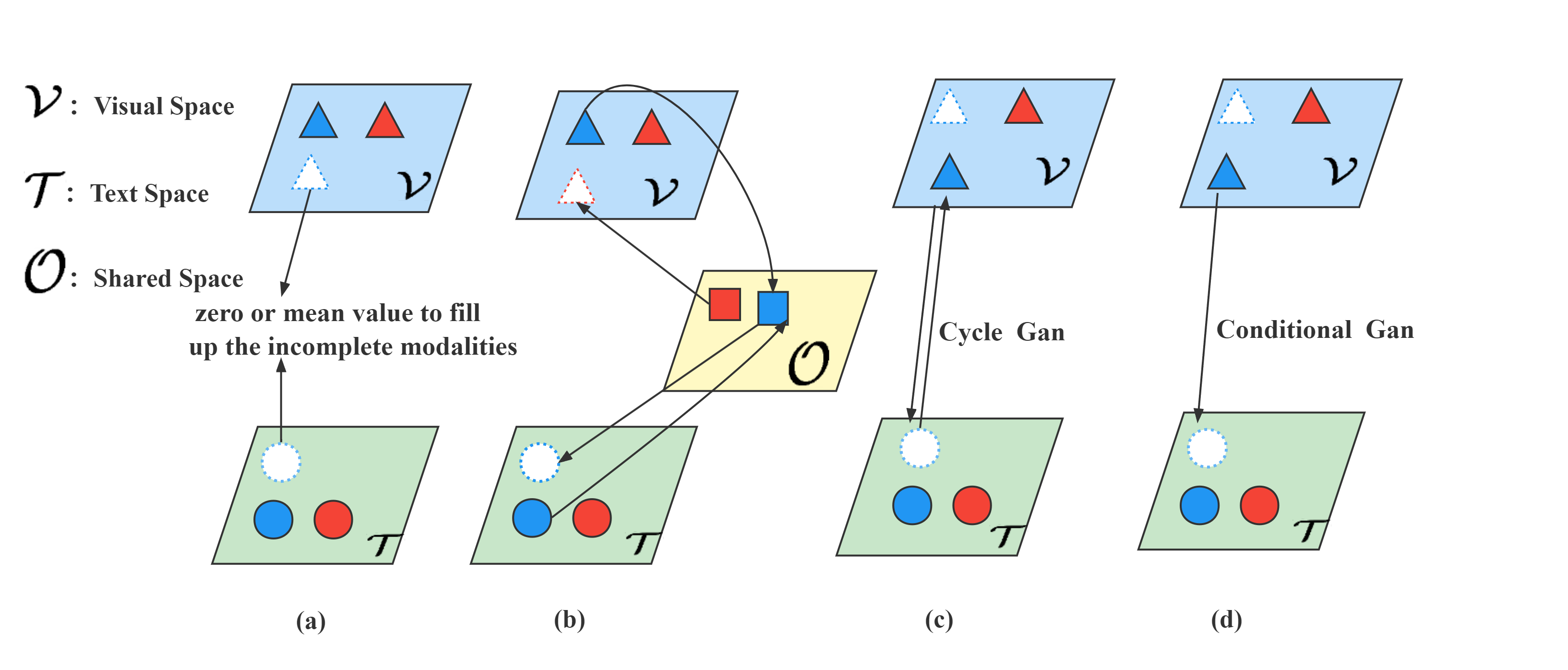}
  \caption{Four imputation  strategies for missing values. (a) Zero or mean-based methods. (b) Common space learning-based method. (c) Cycle GAN-based method. (d) Ours proposed CIGIT-C. Because of their heterogeneity, the solid triangles and circles in the figure represent the image and text modalities. Circles and triangles of the same colour represent the same instance. The graph drawn in the dashed box is a representation of the missing modalities. }\label{paradigms}
  \end{center}
\end{figure*}

However, in real-life cases, some instances could lack paired instances of different modalities. For instance, on Twitter, tweets can contain texts or images, but not all contain paired modalities.
The missing modalities not only lead to information loss but also increase the difficulties of excavating complementary information. In general, most present IITC algorithms rely heavily on the idea of incomplete multi-view clustering. Over the past few years, several efforts have been dedicated to addressing IITC, with various imputation strategies (i.e., zero or mean-based methods, common space learning-based methods, generative methods) to fill in the missing samples. As shown in Figure\ref{paradigms}(a), these methods \cite{wen2021unified} adopt zero or mean values to pad missing data  firstly and then design specific machine learning techniques to conduct multi-view clustering. These simply padded data are very different from the actual data and can reduce the performance of the final clustering.  To alleviate this challenge, a common space learning-based method has been introduced, as shown in Figure\ref{paradigms}(b). The goal of the method is to learn a common representation using complete and aligned modalities and then use the common representation to complete the representation of missing modalities. Such as the kernel \cite{shao2013clustering} \cite{liu2018late} \cite{liu2020efficient} method, the matrix \cite{hu2019one} \cite{rai2016partial} method and the contrastive learning \cite{lin2021completer} method. However, these methods cannot explicitly compensate for the missing data in each modality. To solve this issue, generative-based(GAN\cite{tran2017missing}) methods (see Figure\ref{paradigms}(c)  and \ref{paradigms}(d)) are proposed. As shown in the Figure\ref{paradigms}(c), the third imputation strategy for missing values is based on  Cycle GAN \cite{zhu2017unpaired}.
To tackle a more complex problem when each instance is represented in only one modality, for instance, TPIT-C \cite{guo2022two} integrates the adversarial learning framework approach.

Although the aforementioned methods provide some schemes to address the IITC problem, these methods still suffer from the following issues: 
1) The majority of previously incomplete multi-view clustering techniques extract distinct view features from various visual datasets that are fundamentally homogenous using various feature description techniques (e.g., SIFT, LBP, HOG  \cite{routray2017analysis}).
Therefore, as shown in  Figure \ref{Fig1}, the direct use of these
methods on heterogeneous data (i.e. visual and textual data) may
cause adverse effects and even unsuccessful clustering tasks. This is since they neglect distinct characteristics of visual and textual patterns. The reason for this since images and text each have a different signal space. Language tasks have a discrete signal space (words, sub-word units, etc.). In contrast, the original signals of images are in a continuous, high-dimensional space and are not constructed for human communication (e.g., unlike words).
2) The inaccurate imputation or padding for missing data harms clustering performance. In addition, the missing data processing tends to lean towards the generation task rather than the clustering task. Therefore, the performance of missing value completion and inversion is poor.
3) Existing methods ignore tapping into the latent connections within and between modalities. 

To address the above issues, we propose a novel model named 
CIGIT-C, which first maps the original features of the image and 
text into their respective subspaces using their specific encoders. The different modalities provide cluster-level distinctive representations. This step takes into account the intra-modal connections. Second, inspired by conditional GAN \cite{mirza2014conditional} , as  shown in Figure \ref{paradigms}(d), the two generators use conditional information (cluster label knowledge) provided by their respective subspaces to standardize the generation process to generate representations,  which is beneficial to the clustering results. Clustered adversarial networks can not only generate representations to compensate for missing data. Moreover, these generated representations are beneficial for clustering tasks. The generators thoroughly learn the knowledge of clustering-level across modalities, exploiting and fully exploiting the complementarity between modalities. KL clustering loss is used to update the encoder, which attempts to make the distribution of the learned representations consistent and compact. 
The main contributions of our approach are listed below:

(a) We have argued theoretically (in Section \ref{introduction}) and experimentally (in Section \ref{experiments}) that existing incomplete multi-view methods do not solve the IITC task well. Such a theoretical view is remarkably different from existing works, which treat image-text as homogeneous data. 

 (b) For the processing of missing data, we are driven by clustering tasks rather than generating tasks, which are ignored in the existing IITC methods.

(c) We fully explore the potential relationships within and between modalities. Extensive experiments demonstrate that our method 
achieves superior clustering performance compared to state-of-the-art methods.

\section{RELATED  WORK}\label{sec2}

In this section, we briefly review recent advances in two related topics, namely incomplete multi-view clustering and knowledge distillation.

\subsection{Imcomplete Multi-view Clustering}

Existing works can be roughly divided into three categories based on the imputation strategy.

a) In the pioneering works, imcomplete multi-view clustering (IMVC) uses zero or mean \cite{wen2021unified} values to fill in the missing values. However, these simple padded data are very different from the real data, which avoids learning a consistent clustering structure and severely degrades the final clustering performance. 

b) The second padding strategy for missing values is establishing a common latent subspace using the complete data and then padding the missing data with the learned common representation, i.e., matrix factorization (MF) based IMVC, kernel-based methods, contrastive learning based IMVC. The MF-based approach focuses on obtaining a consensus representation shared by all views from the original feature space. For example, DAIMC \cite{hu2019doubly} establishes a consensus basis matrix with the help of ${\ell_{2,1^{-}}}norm$
. IMG \cite{zhao2016incomplete} employs the graph laplacian
to regularize the latent subspaces of each data view. 
 Hu et al. \cite{hu2019one} propose an efficient and effective method for handling large-scale incomplete multi-view clustering problems using regularized matrix decomposition and weighted matrix factorization, taking full account of the missing information of the instances.
The kernel-based method of main idea are to leverage the kernel 
matrices of the complete views for completing the kernel 
matrix of incomplete view. The first partial multi-modal clustering algorithm is proposed by Rai et al. \cite{rai2010multiview}. To fill the missing modality kernel matrix, they employ a modality kernel representation and the Laplace regularisation approach. Liu et al. \cite{8611131} propose a joint optimization of kernel imputation and clustering that seamlessly integrates these two learning processes to achieve better clustering. EE-IMVC \cite{9001210} incorporates prior knowledge to regularize the learning of the consensus clustering matrix. The motivation for contrastive learning \cite{hadsell2006dimensionality} is to  maximize the similarity of positive pairs and distance of negative pairs. Motivated from the information theory, Lin et al. \cite{lin2021completer} propose a unified framework to jointly learn consistent representation and recover the missing view by maximizing the mutual information while minimizing the conditional entropy of multiple views. Additionally, it shows promising results when applied to complete multi-view data. However, this imputation strategy focuses on the available information of present views. PMVC \cite{li2014partial}, IMG\cite{zhao2016incomplete}, MIC \cite{shao2015multiple}, and DAIMC are all developed as joint partial alignment or weighted matrix decomposition models to investigate data on the alignment complementarity of current viewpoints and reach consensus. They generally use some regularization or add some restrictions to the new representation. The second imputation strategy focuses on available information about present views without attempting to generate missing information related to the missing views. However, it is challenging to discover correlations across all the various modalities when there is a high percentage of missing data.

c) The third  imputation  strategy is to use the generative method GAN to fill in the missing views.
The GAN model is intended to discover the connection between various modalities. The model
from references \cite{wang2018partial} was used to complete the missing data.
Cycle GAN \cite{zhu2017unpaired} uses generative methods and their inverse directions to achieve unpaired image style transformation. Inspired by this, 
Guo et al. \cite{guo2022two} suggest an efficient approach for extracting the global representation of visual instances. Furthermore, this method integrates the adversarial learning framework methodology to handle a more difficult job in which each instance is represented in just one modality. In addition to the above three imputation strategies, some graph-based learning methods \cite{wen2021structural}\cite{xue2021clustering} are
proposed to handle the arbitrary IMC cases. These work fill in missing values by extracting currently available multi-view information, i.e., data structures. CDIMC-net \cite{ijcai2020-447} incorporates the view-specific encoders and the graph embedding strategy to handle the incomplete
multi-view data. However, obtaining the complete data structure is difficult when some views are missing.

We have borrowed some ideas from work above, namely the use of generative models, and the idea is natural to use GAN to fill in the missing values.
The differences between this study and existing works are given below.

1) CIGIT-C uses clustering labels as a supervision to guide the generation process, allowing the generation process to be regularized. CIGIT-C can guarantee that the generative representation is suitable for the clustering task.

2) Instead of using the raw feature space, we apply the generative technique to the latent subspace. This promises to investigate the data's structure and produce more distinctive feature representations.

\subsection{Knowledge Distillation}

Knowledge distillation \cite{bucilu2006model} is a compression technique in which a compact model (student) is trained to reproduce the behaviour of a larger model (teacher) or set of models.

Distilling models \cite{zhai2020multiple} \cite{zhang2018deep} convey extra information beyond the traditional supervised learning target by mimicking the teacher's class probability or feature representation. Recently, in order to address the inconsistency between training and inference caused by the introduction of dropout, Liang et al. \cite{NEURIPS2021_5a66b920} introduce a simple consistent training strategy that normalizes dropout.  As some modalities in the IITC task might be inherent in low quality or inconsistent cross-modality clustering prediction, they will negatively affect the fusion process, in turn, the clustering result. Motivated by these, we design a loss function that minimizes KL-divergence between the clustering prediction scores of the respective modalities and fused representations clustering probability.

\begin{figure*}[t]
  \begin{center}
  \includegraphics[width=0.90\textwidth]{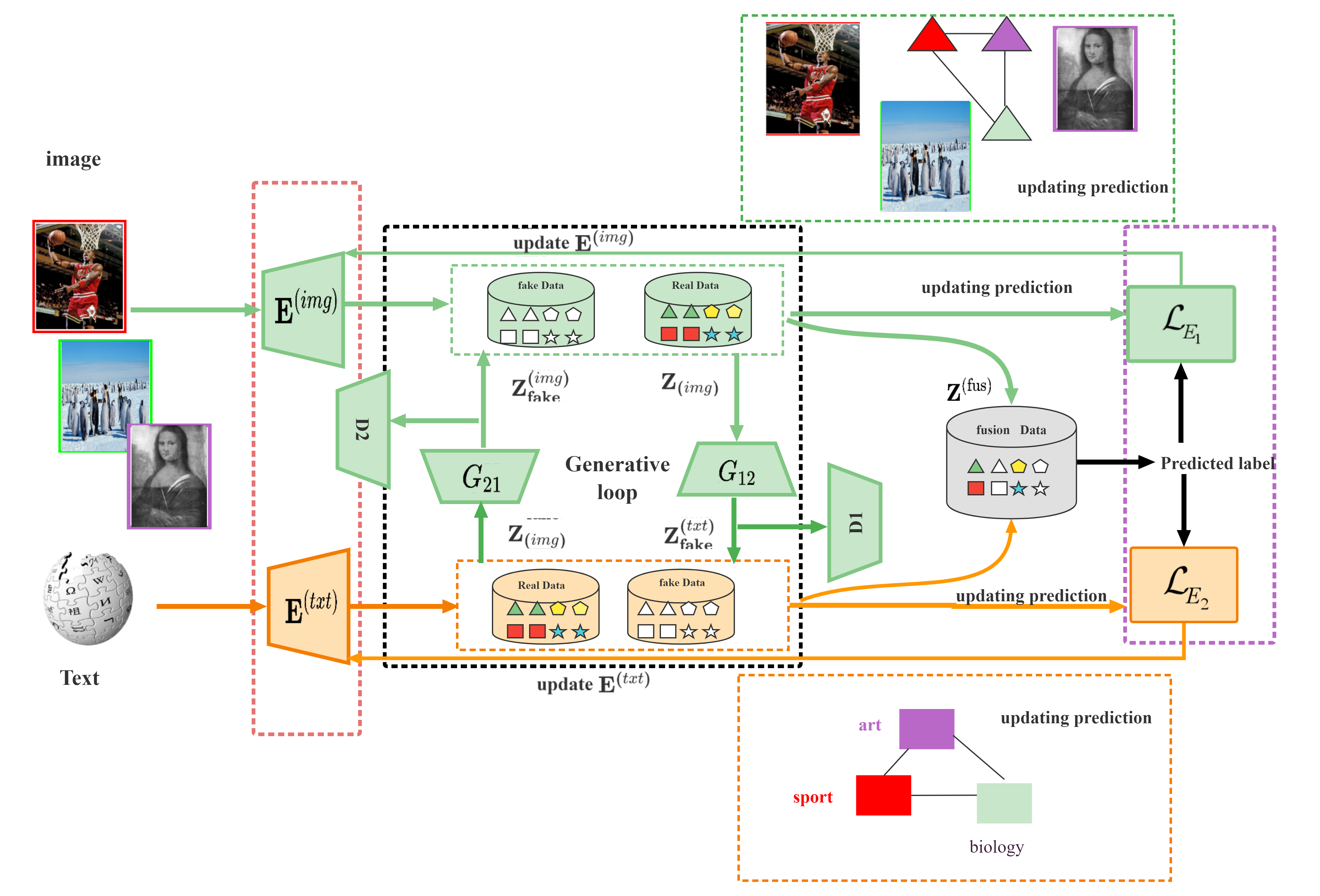}
  \caption{ The framework of our proposed Clustering-Induced Generative Incomplete Image-Text Clustering(CIGIT-C). The model contains two core components, namely modality-specific encoders and subspace conditional clustering GAN, as shown in Figure with the red and black dashed boxes. The role of the modality-specific encoders is to map the original features into the subspace with the aim of seeking distinctive representations in the subspace. The subspace conditional clustering GAN  function is to increase the cross-modal representation diversity and improve the model robustness. These generated representations are beneficial for clustering tasks. In addition, we introduce KL-divergence loss, the purple dashed box in the figure,  making the latent distributions for all the modalities consistent and compact cluster-wise. }\label{circuit_diagram}
  \end{center}
\end{figure*}

\begin{table}[H]
\setlength{\tabcolsep}{1.8mm}{
\caption{ Notations and descriptions}
\begin{tabular}{l|l}
\hline
\textbf{Notations} & \textbf{Descriptions} \\ \hline
${{\mathbf{X}}^{(img)}}$, ${{\mathbf{X}}^{(txt)}}$ & \textbf{Image/Text original features.} \\ \hline
${\widetilde{\mathbf{X}}^{(img)}}$, ${\widetilde{\mathbf{X}}^{(txt)}}$ & \textbf{Text/Image features of data missing modalities.} \\ \hline
${\mathbf{Z}^{(img)}}$,${\mathbf{Z}^{(txt)}}$
&
\textbf{Subspace representation of images/text.}  \\ \hline
${\mathbf{E}^{(img)}}$,${\mathbf{E}^{(txt)}}$ &
\textbf{Encoders of image/text modalities.} \\ \hline ${\mathbf{Z}_{\text {fake }}^{(img)}}$,${\mathbf{Z}_{\text {fake }}^{(txt)}}$
&\textbf{Generated fake images/text representations.} \\ \hline
${\mathbf{Z}^{\text{(fusion)}}}$
&\textbf{Fusion representations} \\

\hline
\end{tabular}}
\end{table}
% Please add the following required packages to your document preamble:
% \usepackage[table,xcdraw]{xcolor}
% If you use beamer only pass "xcolor=table" option, i.e. \documentclass[xcolor=table]{beamer}

\begin{figure}[h]
\centering
\includegraphics[width=0.5\textwidth]{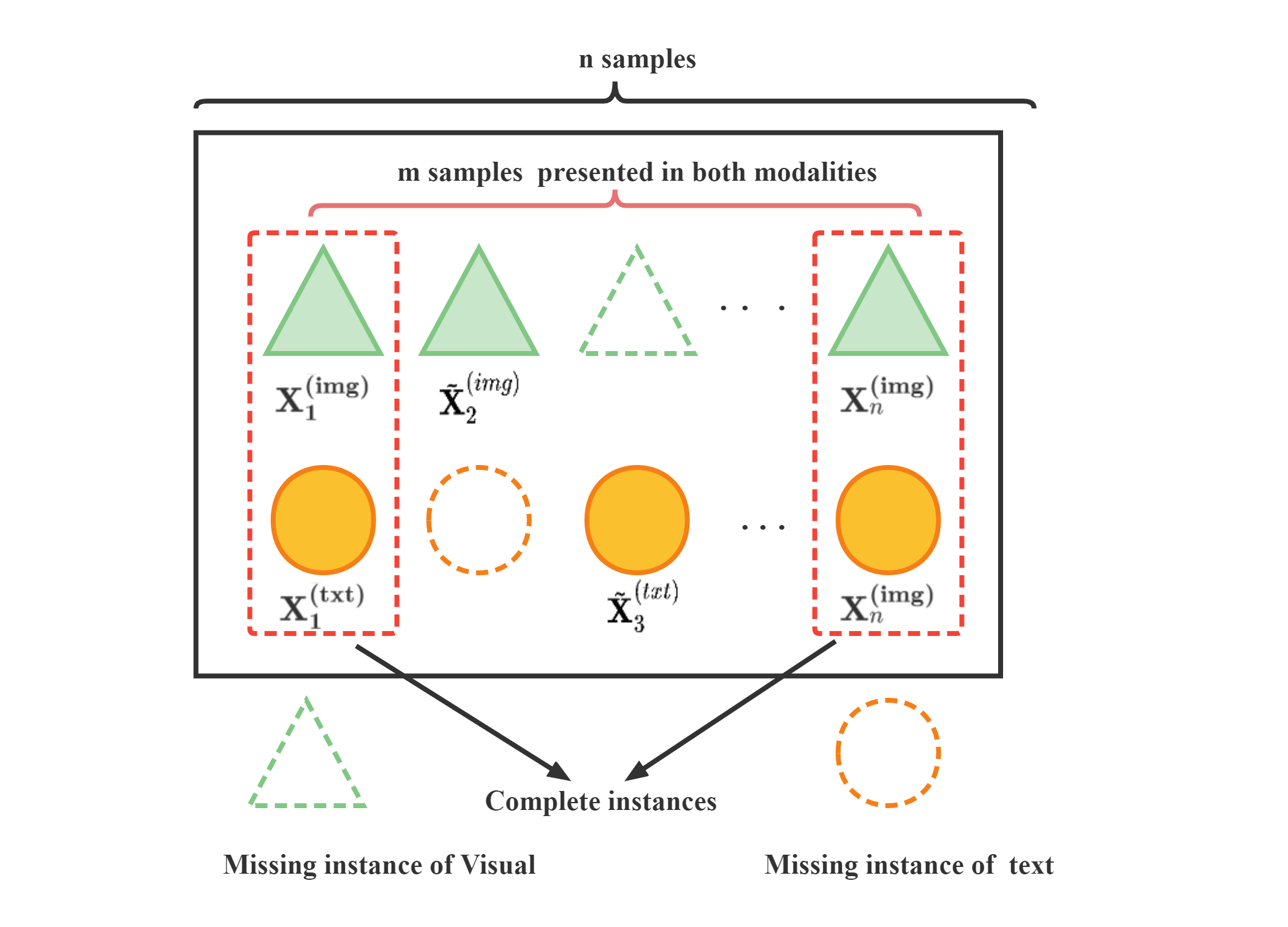}
\caption{Notions for incomplete image-text data. The green triangle and the orange circle represent the image and text modalities, respectively. The red dashed boxes ${\mathbf{X}_{n}^{(\text {img) }}}$ and ${\mathbf{X}_{n}^{(\text {txt) }}}$ indicate the same instance, respectively, and the samples' modalities were accessible and strictly aligned. ${\widetilde{\mathbf{X}}^{(img)}}$ and ${\widetilde{\mathbf{X}}^{(txt)}}$ denote the image modality only, and the text modality only, respectively. ${n}$ denotes the number of data points of the whole dataset. ${m}$ denotes the number of examples presented in all modalities.}
\end{figure}

\section{CIGIT-C: THE PROPOSED MODEL}\label{sec3}
In this section, we first introduce the notations and descriptions. After clarifying this,  we present an overview of the CIGIT-C model. Since CIGIT-C consists of three components, we describe each component and the loss function in detail. Then, we provide a method to optimize the CIGIT-C model. Finally, we integrate the above to provide a discussion of the CIGIT-C model.

\subsection{Notations and Descriptions}
In the legacy ITC task, each sample contains features in all the 
modalities. This condition, however, cannot always be held in the 
incomplete ITC task. Given a dataset  ${\mathcal{D}=\left\{{\mathbf{X}}^{(img,txt)},  \tilde{\mathbf{X}}^{(img)},\tilde{\mathbf{X}}^{(txt)}\right\}}$ of n instances, where ${{\mathbf{X}}^{(img,txt)}}$,  ${\tilde{\mathbf{X}}^{(img)}}$, ${\tilde{\mathbf{X}}^{(txt)}}$ denote the examples presented in both modalities, the visual modality only, and the text modality only, respectively. For clarity, we denote
variables originated from images and texts by superscript img and text,
respectively. The frequently used notations and their descriptions are listed in Table 1. Because complete modal data are few and expensive, it is advantageous to fully use the substantial amount of incomplete-modal data.
\subsection{Model Overview}
The original features of the image and text are first mapped to a more distinctive subspace through encoders ${\mathbf{E}^{(img)}}$ and ${\mathbf{E}^{(txt)}}$, respectively. More distinctive representations were found by subspace clustering. The two generators generate a clustering task-driven representation based on the clustering information provided by the other subspace. The generative mechanism fully uses the characteristics and cluster distribution of ${\mathbf{Z}^{(img)}}$ and ${\mathbf{Z}^{(txt)}}$. The generated representation is fused with the original representation for clustering. Furthermore, at the same time, aligned clustering loss is designed to obtain a better modality-specific encoder.
\subsection{Modality-Specific Encoders}
The role of the modality-specific encoders ${\operatorname{E}^{(img)}}$ and ${\operatorname{E}^{(txt)}}$ is to map the original features into a more distinctive subspace. The reason for this is twofold: 1) Given that the frequencies, formats and receptive domains of images and text are very different. Furthermore, since there may be higher-order noise in the original features, directly applying GAN to the original features would hurt the clustering predictions and make GAN harder to train and converge. 2) Different 
modalities can provide uniqueness at different clustering levels, and exploring the correlation between different categories and modalities is critical to improving the clustering performance further. The clustering label information obtained from the subspace is also used to standardise the generation of Subspace Conditional Clustering GAN to benefit the features of the clustering task.

\begin{equation}
\label{Eq1}
Z_{n}^{(fus)}=\sum_{t=1}^{n}(1-\beta)E^{(img)}(\mathbf{X}_{t}^{img})+\beta E^{(txt)}(\mathbf{X}_{t}^{txt})
\end{equation}
${\operatorname{E}^{(img)}\left(\boldsymbol{x}^{img} ; \theta^{img}\right)}$ and ${\operatorname{E}^{(txt)}\left(\boldsymbol{x}^{txt} ; \theta^{txt}\right)}$ in Eq \ref{Eq1} represent the encoders for images and text, respectively, where ${\theta}$ is the parameter that can be learned by back-propagation. ${\boldsymbol{z}_{i}^{img}=E^{img}\left(\boldsymbol{x}_{i}^{img}\right)}$ and ${\boldsymbol{z}_{i}^{txt}=E^{txt}\left(\boldsymbol{x}_{i}^{txt}\right)}$ represent the subspace features, while ${\mathbf{Z}^{(
fus)}}$ is the subspace features of the fused image and text, where ${\beta}$ is a weighting factor to balance the ratio of text and visual modality.
Moreover, in order to make the projected samples more distinctive across modalities, thus, the  Student's t-distribution is utilized, where the goal of Student's t-distribution is to make the projected representations closer to the samples of the same the clustering centre than it is to any other cluster centre. As follow Eq \ref{Eq2}, ${\delta}$ represents the degrees of freedom of the Student's t-distribution. Following Xie et al. \cite{xie2016unsupervised}, we set ${\delta}$ to 1. The clustering centroid of the ${K-th}$ cluster in the subspace for the ${m-th}$ modality is denoted by ${\mu_{k}^{(m)}}$. The centroids are initialized by ${K}$-means and updated by Stochastic Gradient Descent (SGD). 
${\mathbf{z}_{n}^{(m)}}$ represents learning the subspace representation learned through the encoder, where m=1 represents the visual modality ${\mathbf{Z}^{(img)}}$, m=2, represents the text modality ${\mathbf{Z}^{(txt)}}$, and m=3 represents the fusion modality ${\mathbf{Z}^{(fus)}}$. ${q_{n k}^{(m)}}$  represents the probability of assigning instance n to group K of the ${m-th}$ modality (i.e., soft assignment).
\begin{equation}
\label{Eq2}
q_{n k}^{(m)}=\frac{\left(1+\left\|\mathbf{z}_{n}^{(m)}-\mu_{k}^{(m)}\right\|_{2}^{2} / \delta\right)^{-\frac{\delta+1}{2}}}{\sum_{k^{\prime}}\left(1+\left\|\mathbf{z}_{n}^{(m)}-\mu_{k^{\prime}}^{(m)}\right\|_{2}^{2} / \delta\right)^{-\frac{\delta+1}{2}}}
\end{equation}
To enhance cluster compactness, we prioritise data points assigned with high confidence by calculating the emphasised target distribution ${p_{n k}}$ as follows:
\begin{equation}
    p_{n k}=\frac{q_{n k}^{2} / f_{k}}{\sum_{k^{\prime}} q_{n k^{\prime}}^{2} / f_{k^{\prime}}}
    \label{eq3}
\end{equation}
where ${f_{k}=\sum_{n} q_{n k}}$ denotes soft cluster frequencies. Before normalising it by frequency per cluster, it first squares ${q_{n k}}$. Thus, emphasis will be placed on high probability.
Thus the modality-specific encoder is updated by the following loss ${\mathcal{L}_{E_{m}}}$, as follows:
\begin{equation}
\begin{aligned}
\mathcal{L}_{E_{m}} &=K L\left(P^{(m)} \| Q^{(m)}\right)+\alpha K L\left(P^{(3)} \| Q^{(3)}\right) \\
&=\sum_{n} \sum_{j} p_{n k}^{(m)} \log \frac{p_{n k}^{(m)}}{q_{n k}^{(m)}}+\alpha \sum_{n} \sum_{j} p_{n k}^{(3)} \log \frac{p_{n k}^{(3)}}{q_{n k}^{(3)}},
\end{aligned}
\label{eq4}
\end{equation}
where m=1 and m=2 correspond to the losses of encoders ${\operatorname{E}^{(img)}}$ and ${\operatorname{E}^{(txt)}}$, respectively, and ${\alpha}$ refers to a trade-off parameter. The objective ${\mathcal{L}_{E_{m}}}$ is to match the soft assignment ${p_{n k}}$ with the desired distribution  ${q_{n k}}$. Loss ${\mathcal{L}_{E_{m}}}$ is designed to resolve inconsistencies in cluster predictions for different specific modalities. Loss ${\mathcal{L}_{E_{m}}}$ forces the distribution of fused features to be consistent with that of different modalities. Specifically, minimize the KL-divergence of fusing feature  with particular 
modality image and text.

\begin{figure}[h]
\centering
\includegraphics[width=0.5\textwidth]{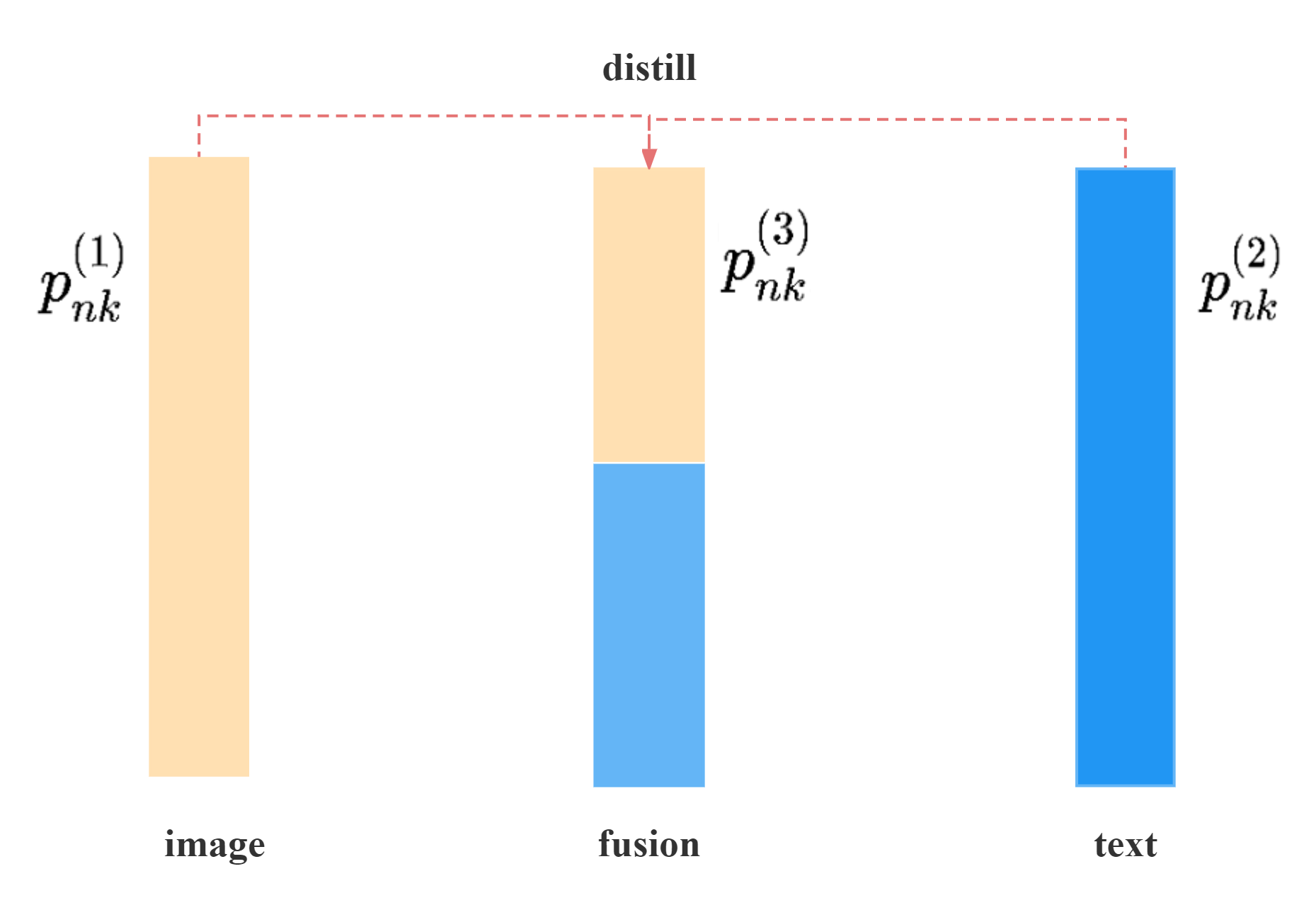}
\caption{${ p_{n k}^{(1)}}$ and ${ p_{n k}^{(2)}}$  represent the logits scores of the respective modalities, while ${ p_{n k}^{(3)}}$  represents the fused logits scores. We use KL-divergence to add consistency constraints, i.e. different modalities and fused modalities predict the clustering knowledge consistently. This learns a common semantic representation of the visual and text.}
\end{figure}

\subsection{Subspace Conditional Clustering GAN}

Subspace conditional clustering GAN module contains two
 generators, ${G_{12}}$ and ${G_{21}}$, and their corresponding discriminators, ${D_{1}}$ and ${D_{2}}$, which are trained in the inverse direction. The generator is trained to generate samples while the discriminator attempts to differentiate the samples. Both networks are compelled to improve their capabilities through a competition strategy. Since ${G_{12}}$, ${D_{1}}$, and ${G_{21}}$, ${D_{2}}$, have symmetrical positions and identical objective equations. So here, we will only discuss G12 and D1.
 
 \begin{equation}
 \begin{aligned}
     L_{G_{12} d}&=-E_{z \sim p_{z}(z)} \log  \Big(1-D_{1}\Big(G_{12} 
     \Big(z \mid E^{(img)}\Big(X_{t}^{(img)}\Big)\\ &\Big)\Big)\Big),
     \label{eq5}
     \end{aligned}
 \end{equation}
where ${z}$ refers to noise, subspace conditional clustering GAN differs from traditional GAN in focusing on clustering rather than generation tasks. ${z}$ is an a prior sampling method consisting of a cascade of normal random variables and a one-hot label noise.
Since the subspaces ${\mathbf{Z}^{(img)}}$ and ${\mathbf{Z}^{(txt)}}$ are modified when the encoders ${\mathbf{E}^{(img)}}$ and ${\mathbf{E}^{(txt)}}$ are tuned, it is challenging to achieve stable generative outputs directly. Therefore, we introduce a similarity constraint that pulls produced samples and real samples toward subspace similarity. The objective phrase is depicted below:
\begin{small}
\begin{equation}
\begin{aligned}
L_{G_{12} s}&=E_{z \sim p_{z}(z)}\Big(\Big\|G_{12}\left(z \mid E^{(img)}\left(X^{img}\right)\right) \\ 
&-E^{(txt)}\left(X^{txt}\right)\Big\|_{\mathrm{F}}^{2}\Big)
\label{eq6}
\end{aligned}
\end{equation}
\end{small}

The final overall objective ${L_{G_{12}}}$ is obtained by combining Eq.\ref{eq5} and \ref{eq6} , as shown in Eq.\ref{eq7} below, where ${\mu}$ is used to balance discriminator loss and similarity loss.

\begin{equation}
  L_{G_{12}}= L_{G_{12} d} +\mu  L_{G_{12} s}
 \label{eq7}
\end{equation}
The function of ${D_{1}}$ is to distinguish the generated samples as much as possible from the real samples in subspace ${\mathbf{Z}^{(txt)}}$.
\begin{equation}
 \begin{aligned}
L_{D_{1}} &=E_{X \sim p_{X}(X)} \log D_{1}\left(E^{(txt)}\left(X^{txt}\right)\right) \\
&+E_{z \sim p_{z}(z)} \log \left(1-D_{1}\left(G_{12}\left(z \mid E^{(img)}\left(X^{img}\right)\right)\right)\right)
\end{aligned}
\label{eq8}
\end{equation}
After the representation is generated in the subspace of modality-specific, the false representation and the true representation are fused, and  Eq.\ref{Eq1} is updated as:
\begin{equation}
     \begin{aligned}
Z_{n}^{(fus)}=&\sum_{t=1}^{n}(1-\beta)E^{(img)}(\mathbf{X}_{t}^{img})\\&+\beta E^{(txt)}(\mathbf{X}_{t}^{txt})
+\sum_{m=1}^{2} \eta _{m} Z_{\text {fake }}^{(m)},
\end{aligned}
\label{eq9}
\end{equation}
where ${\eta _{m}}$ is a trade-off parameter that balances the weight between the real representation and the generated representation, m=1 represents the visual modality, while m=2 represents the text modality.

Finally, we define the overall loss function
of CIGIT-C as:

\begin{equation}
    \mathcal{L}_{\text {total}}=\min _{E_{m}, G_{12},G_{21}} \max _{D_{1},D_{2}} \mathcal{L}_{E_{m}}+\mathcal{L}_{G_{12}}+\mathcal{L}_{G_{2 1}}+\mathcal{L}_{D_{1}}+\mathcal{L}_{D_{2}}.
\end{equation}
For a clear understanding, we summarize the optimization steps of the proposed method CIGIT-C in Algorithm \ref{Algorithm 1}.

\subsection{Model Analysis and  Learning}
Our model, CIGIT-C, differs from other approaches in three main ways: 1) The generation mechanism focuses on clustering tasks rather than generating tasks, ensuring that the generated representation is beneficial to the task. 2) The CIGIT-C model taps into latent relationships within and between modalities, on the one hand, mapping raw features to more distinctive subspaces aiming to allow different modalities to provide cluster-level unique representations. This aspect considers intra-modal connections. On the other hand, adversarial learning is encapsulated in the CIGIT-C model to explore complementary information across modalities. This aspect takes into account the inter-modal connections.
In addition, subspace conditional clustering GAN is also used as regularization to enhance the robustness and accuracy of the model.
3) We consider and resolve inconsistent specific-modality clustering predictions. We introduce the effect of knowledge distillation loss, on the one hand, to make the posterior probability of the fused feature consistent with the posterior probability of each modality subspace, and on the other hand, to enable the encoder to better extract the semantic knowledge within the subspace and use it in the representation of the embedded subspace.

\begin{algorithm}[htb]
\caption{Clustering-Induced Generative Incomplete
Image-Text Clustering (CIGIT-C) Algorithm.}
\label{Algorithm 1}
\begin{algorithmic}[1] %这个1 表示每一行都显示数字
\REQUIRE ~~\\ %算法的输入参数：Input
    ${\mathcal{D}=\left\{{\mathbf{X}}^{(img,txt)},  \tilde{\mathbf{X}}^{(img)},\tilde{\mathbf{X}}^{(txt)}\right\}}$, where some samples either lack image or text modality; learing rate $\omega$; hyper-parameter  $\beta,\alpha,\eta_{1},\eta _{2}$; iteration number ${I}$; cluster number ${K}$.
\ENSURE ~~\\ %算法的输出：Output 
   The optimal parameters ${\Theta_{G_{12}}, \Theta_{G_{21}}, \Theta_{D_{1}}, }$
and ${\Theta_{D_{2}}}$; clustering results $\mathbf{R}$;
    \STATE 
    Map original features ${\left\{{\mathbf{X}}^{(img,txt)},  \tilde{\mathbf{X}}^{(img)}\right\}}$ to more distinctive subspaces ${\left\{{\mathbf{Z}^{(img)},\mathbf{Z}^{(txt)} }\right\}}$. Initialize the parameters of CIGIT-C with Xavier initializer. Calculate fusion features ${\mathbf{Z}^{\text{(fusion)}}}$ and clustering centres  ${\mu_{k}^{(m)}}$. ${m \in\{1=img,2=txt,3=fus\}}$.

     \WHILE {${t \leq \text { MaxIter }}$}
        \STATE Train the encoders with ${\mathcal{L}_{E_{m}}}$ according to Eq.\ref{eq4}. ${\forall m=1,2}$
        \STATE Train generators ${G_{12}}$ and ${G_{21}}$ using loss functions ${L_{G_{12}}}$ and ${L_{G_{21}}}$ according to Eq. \ref{eq7}.
        \STATE Train discriminators ${D_{1}}$ and ${D_{2}}$ using loss functions ${L_{D_{1}}}$ and ${L_{D_{2}}}$ according to Eq. \ref{eq8}.
         \STATE The fusion features are updated according to Eq.\ref{eq9}, and ${\mu_{k}^{(m)}}$ is recalculated. ${m \in\{1=img,2=txt,3=fus\}}$.
    \ENDWHILE 
    
    \STATE The final updated fusion representation ${{\mathbf{Z}^{\text{(fusion)}}}}$ is obtained, and ${\mu_{k}^{(3)}}$ and ${q_{n k}^{(3)}}$ are computed in turn by using Eq.\ref{Eq2} and \ref{eq3}, respectively.
    \STATE Obtain the clustering results $\mathbf{R}$;
\RETURN $\mathbf{R}$; %算法的返回值
\end{algorithmic}
\end{algorithm}
\section{EXPERIMENTS}
\label{experiments}
\subsection{Experimental settings}

All tests are carried out on the Ubuntu 16.04 platform, with Tesla P100 Graphics Processing Units and 32GB of RAM. The PyTorch 1.6.0 framework serves as the foundation for our model, technique, and baseline.
\subsubsection{Compared methods}
\

We use state-of-the-art methods to demonstrate the effectiveness of our model CIGIT-C. Comparison baselines are briefly introduced below.
\textbf{BestSM}:
We report the best
clustering results achieved by performing k-means on modality-specific, i.e., visual features ${{\mathbf{X}}^{(img)}}$
and text features  ${{\mathbf{X}}^{(txt)}}$.
\textbf{DAIMC\cite{ijcai2018-313}}:
It is based on weighted semi-nonnegative matrix factorization to
obtain cluster results. \textbf{UEAF\cite{wen2019unified}}: A locality-preserved reconstruction term is introduced to infer the missing views so that all views can be aligned.
\textbf{OPIMC\cite{hu2019one}}:  It adopts regularized and weighted matrix
factorization to obtain clustering results.
\textbf{PIC\cite{wang2019spectral}}:  It learns a consensus Laplacian matrix from incomplete multi-view data for clustering. 
\textbf{CDIMC-net\cite{ijcai2020-447}}:
It incorporates the view-specific encoders and the graph embedding
strategy to handle the incomplete multi-view data.
\textbf{CASEN\cite{xue2021clustering}}:
It is an end-to-end trainable framework that jointly conducts multi-view structure enhancing and data clustering.
\textbf{COMPLETER\cite{lin2021completer}}: It is a two-stage contrastive multi-view clustering method by learning consistent features. 

\subsubsection{Datasets and Features}
\ 

\textbf{Wikipedia} \cite{pereira2013role} contains 2,866 image-text pairs, including ten semantic categories, sourced from the Wikipedia collection of "featured articles". The text in each pair is a paragraph describing the content of the corresponding image.
For the test set, we chose 30${\%}$ randomly from the dataset. The result forms a training set of 1910 samples and a test set with 956 samples.
\textbf{ NUS-WIDE-10k} consists of 10,000 image-text pairs, selected on average from the ten largest semantic categories of the NUS-WIDE \cite{chua2009nus} dataset.
The 10,000 selected image-text pairs were divided into two parts: 8,000 image-text pairs for training and 2,000 image-text pairs for testing.

\textbf{BDGP} \cite{cai2012joint} contains 2500 instances about drosophila embryos of 5 categories.

For all subsequent comparisons, we use the same image and text features for a fair comparison. For the BDGP dataset, the feature files are provided by the authors.
For the image samples of Wikipedia and NUS-WIDE-10K, we first resize 
it into 224${\times}$224 then employ the VGGNet-19 \cite{simonyan2014very} model pre-trained on the ImageNet to output 4096-dimensional features from its fc\emph{7} layer. For the text samples of Wikipedia, we adopt a pre-trained BERT to extract 768-dimensional text features. Similar to the BERT paper \cite{devlin2018bert}, we take the
embedding associated with ${[CLS]}$ to represent the whole sentence.
For the text samples of NUS-WIDE-10K, we adopt the 
widely-used bag-of-words(BoW) vector with the TF-IDF weighting 
approach  to get 1000-dimensional features. The detailed statistics of the three datasets are summarized in Table \ref{Table2}.

\subsubsection{Incomplete data construction}
\

To demonstrate the effectiveness of the CIGIT-C model in processing missing image-text data. We set different missing rates ${p\in\{10\%,30\%,50\%,70\%\}}$. The missing rate ${p}$ is defined as ${p=(n-m)/ n}$, where m is the 
number of complete examples, and n is the number of the 
whole dataset. We randomly remove a modality by the missing rate ${p}$ to select some instances as incomplete data. A high missing rate is positively correlated with task complexity.

\begin{table}[H]
\caption{The statistics of the datasets.}
\label{Table2}
\setlength{\tabcolsep}{1.8mm}{
\begin{tabular}{lllll}
\hline \hline
Datasets & Classes & Modality & Samples & Feature \\ \hline
Wikipedia & 10 & \begin{tabular}[c]{@{}l@{}}Image\\ Text\end{tabular} & \begin{tabular}[c]{@{}l@{}}1919/956\\ 1919/956\end{tabular} & \begin{tabular}[c]{@{}l@{}}4096D VGG\\ 768d BERT\end{tabular} \\ \hline
NUS-WIDE-10K & 10 & \begin{tabular}[c]{@{}l@{}}Image\\ Text\end{tabular} & \begin{tabular}[c]{@{}l@{}}8000/2000\\ 8000/2000\end{tabular} & \begin{tabular}[c]{@{}l@{}}4096D VGG\\ 1000d  BoW\end{tabular} \\ \hline
BDGP & 5 & \begin{tabular}[c]{@{}l@{}}Image\\ Text\end{tabular} & \begin{tabular}[c]{@{}l@{}}2000/500\\ 2000/500\end{tabular} & \begin{tabular}[c]{@{}l@{}}1000D lateral\\ 79d texture\end{tabular} \\ \hline \hline
\end{tabular}}
\end{table}
\subsubsection{Evaluation Metric}
\

We choose two classical and widely used clustering metrics to measure the consistency of cluster assignments and  ground-truth memberships:

 (1) Clustering accuracy (ACC) maps one-to-one the learned clusters to the
ground-truth classes by the Hungarian algorithm and measures the classification accuracy; 

(2) Normalised mutual information (NMI) quantifies the labelling consistency
by the normalised MI between the predicted and ground truth labels of all samples.

For all the metrics, the value 1 means a perfect clustering.

\subsubsection{Implementation details}

\begin{table*}[h]
\centering
\caption{The architecture of the encoders in CIGIT-C}
\label{Table3}
\setlength{\tabcolsep}{13mm}{
\begin{tabular}{|lll|}

\hline
Dataset      & ${\mathbf{E}^{(img)}}$                                                                                                                                         & ${\mathbf{E}^{(txt)}}$                                                              \\ \hline
Wikipedia    & \begin{tabular}[c]{@{}l@{}}Input (size=(bathsize,4096))\\ Dense (LeakyReLU, size =2048)\\  Dense (LeakyReLU, size =1024)\\ Dense (LeakyReLU, size =256) \end{tabular}        & \begin{tabular}[c]{@{}l@{}}Input (size=(bathsize,768))\\ Dense (LeakyReLU, size =256)\\ Dense (LeakyReLU, size =256)\\ Dense (LeakyReLU,size =128)\end{tabular}       \\ \hline
NUS-WIDE-10K & \begin{tabular}[c]{@{}l@{}}Input (size=(bathsize,4096))\\ Dense (LeakyReLU, size =2048) \\ Dense (LeakyReLU, size =1024) \\ Dense (LeakyReLU,size =256))\end{tabular} & \begin{tabular}[c]{@{}l@{}}Input (size=(bathsize,1000))\\  Dense (LeakyReLU, size =256) \\  Dense (LeakyReLU,size =256)) \\ Dense (LeakyReLU,size =128)\end{tabular} \\ \hline
BDGP         & \begin{tabular}[c]{@{}l@{}}Input (size=(bathsize,1000))\\  Dense (LeakyReLU, size =256)\\  Dense (LeakyReLU, size =256)\\ Dense (LeakyReLU,size =128)\end{tabular}    & \begin{tabular}[c]{@{}l@{}}Input (size=(bathsize,79))\\  Dense (LeakyReLU, size =79)\\   Dense (LeakyReLU, size =79)\\  Dense (LeakyReLU,size =64)\end{tabular}     \\ \hline
\end{tabular}}
\end{table*}
\begin{table*}[h]
\centering
\caption{The architecture of the generators in CIGIT-C}
\label{Table4}
\setlength{\tabcolsep}{6mm}{
\begin{tabular}{|lll|}

\hline
Dataset      & ${G_{12}}$                                                                                                                                                                                                              & ${G_{21}}$                                                                                                                                                                                                              \\ \hline
Wikipedia    & \begin{tabular}[c]{@{}l@{}}Input (batchsize,4096+noise-dim=4396)\\ Dense (BatchNorm1d,LeakyReLU, size =1024)\\ Dense (BatchNorm1d, LeakyReLU, size = 1024)\\ Dense (BatchNorm1d,LeakyReLU, size =768)\end{tabular}  & \begin{tabular}[c]{@{}l@{}}Input (batchsize,768+noise-dim=1,068)\\ Dense (BatchNorm1d,LeakyReLU, size =2048)\\ Dense (BatchNorm1d,LeakyReLU, size =2048)\\ Dense (BatchNorm1d,LeakyReLU, size =4096)\end{tabular} \\ \hline
NUS-WIDE-10K & \begin{tabular}[c]{@{}l@{}}Input (batchsize,4096+noise-dim=4396)\\ Dense (BatchNorm1d, LeakyReLU, size = 1024)\\ Dense (BatchNorm1d,LeakyReLU, size =1024)\\ Dense (BatchNorm1d,LeakyReLU, size =1000)\end{tabular} & \begin{tabular}[c]{@{}l@{}}Input (batchsize,1000+noise-dim=1300)\\ Dense (BatchNorm1d,LeakyReLU, size =2048)\\ Dense (BatchNorm1d,LeakyReLU, size =2048)\\ Dense (BatchNorm1d,LeakyReLU, size =4096)\end{tabular} \\ \hline
BDGP         & \begin{tabular}[c]{@{}l@{}}Input (batchsize,1000+noise-dim=1300)\\ Dense (BatchNorm1d,LeakyReLU, size =256)\\ Dense (BatchNorm1d,LeakyReLU, size =128)\\ Dense (BatchNorm1d,LeakyReLU, size =79)\end{tabular}       & \begin{tabular}[c]{@{}l@{}}Input (batchsize,79+noise-dim=379)\\ Dense (BatchNorm1d,LeakyReLU, size =512)\\ Dense (BatchNorm1d,LeakyReLU, size =512)\\ Dense (BatchNorm1d,LeakyReLU, size =1000)\end{tabular}         \\ \hline
\end{tabular}}
\end{table*}

\begin{table*}[h]
\centering
\caption{The architecture of the discriminators in CIGIT-C}
\label{Table5}
\setlength{\tabcolsep}{6mm}{
\begin{tabular}{|lll|}

\hline
Dataset      & ${D_{1}}$                                                                                                                                                                                                                  & ${D_{2}}$                                                                                                                                                                                                                \\ \hline
Wikipedia    & \begin{tabular}[c]{@{}l@{}}Input (batchsize,${\mathbf{Z}_{\text {fake }}^{(txt)}}$+${q_{n k}^{(2)}}$ =768+10=778)\\ Dense (LeakyReLU, size =512)\\ Mini-batch(LeakyReLU, size = 512)\\ Dense (sigmoid, size =1)\end{tabular}  & \begin{tabular}[c]{@{}l@{}}Input (batchsize, ${\mathbf{Z}_{\text {fake }}^{(img)
}}$+${q_{n k}^{(1)}}$ =4096+10=4160))\\ Dense (LeakyReLU, size =1024)\\ Mini-batch(LeakyReLU, size =512)\\ Dense (sigmoid, size =1)\end{tabular} \\ \hline
NUS-WIDE-10K & \begin{tabular}[c]{@{}l@{}}Input (batchsize, ${\mathbf{Z}_{\text {fake }}^{(img)}}$+${q_{n k}^{(2)}}$ =768+10=778)\\ Dense (LeakyReLU, size = 512)\\  Mini-batch(LeakyReLU, size =512)\\ Dense (sigmoid,size =1)\end{tabular} & \begin{tabular}[c]{@{}l@{}}Input (batchsize, ${\mathbf{Z}_{\text {fake }}^{(img)
}}$+${q_{n k}^{(1)}}$ = 4096+10=4160)\\ Dense (LeakyReLU, size =1024)\\ Mini-batch (LeakyReLU, size =512)\\ Dense (sigmoid,size =1)\end{tabular} \\ \hline
BDGP         & \begin{tabular}[c]{@{}l@{}}Input (batchsize,${\mathbf{Z}_{\text {fake }}^{(txt)}}$+${q_{n k}^{(2)}}$ =79+10=89))\\ Dense (LeakyReLU, size =64)\\  Mini-batch (LeakyReLU, size =64)\\ Dense (sigmoid, size =1)\end{tabular}       & \begin{tabular}[c]{@{}l@{}}Input (batchsize, ${\mathbf{Z}_{\text {fake }}^{(img)}}$+${q_{n k}^{(1)}}$ =1000+10=1010)\\ Dense (BatchNorm1d,LeakyReLU, size =512)\\ Mini-batch (BatchNorm1d,LeakyReLU, size =512)\\ Dense (sigmoid,size =1)\end{tabular}         \\ \hline
\end{tabular}}
\end{table*}

\begin{table*}[h]
\centering
\caption{Clustering results (ACC and NMI) of different methods on the Wikipedia dataset with different missing rate ${p\%}$.}
\label{6}
\setlength{\tabcolsep}{6.0mm}{
\begin{tabular}{lllllllll}
\hline
\multicolumn{1}{c}{\multirow{2}{*}{Method}} & \multicolumn{4}{c}{Accuracy}                                                           & \multicolumn{4}{c}{NMI}                                           \\ 
\multicolumn{1}{c}{}                        & 10${\%}$               & 30${\%}$             & 50${\%}$                & 70${\%}$                                     & 10${\%}$               & 30${\%}$                & 50${\%}$                & 70${\%}$              \\ \hline
\multicolumn{1}{l|}{BestSM}                 & 44.79          & 40.70          & 33.45          & \multicolumn{1}{l|}{23.32}          & 48.55          & 39.51          & 31.80           & 22.31          \\
\multicolumn{1}{l|}{DAIMC\cite{ijcai2018-313}}                  & \textbf{55.05}          & 43.70           & 31.96          & \multicolumn{1}{l|}{18.99}          & 45.18          & 28.43          & 16.84          & 9.86           \\
\multicolumn{1}{l|}{UEAF\cite{wen2019unified}}                   & 53.03          & 43.53          & 33.79          & \multicolumn{1}{l|}{25.12}          & 49.83          & 38.12          & 26.64          & 18.59          \\
\multicolumn{1}{l|}{OPIMCC\cite{hu2019one}}                  & 45.15          & 28.28          & 16.92          & \multicolumn{1}{l|}{9.010}           & 54.43          & 39.92          & 33.63          & 25.75          \\
\multicolumn{1}{l|}{PIC\cite{wang2019spectral}}                    & 45.60           & 40.66          & 30.81          & \multicolumn{1}{l|}{26.74}          & 34.42          & 28.84          & 15.02          & 11.37          \\
\multicolumn{1}{l|}{CDIMC-net\cite{ijcai2020-447}}              & 31.28         & 28.45          & 27.43          & \multicolumn{1}{l|}{23.18}          &  36.88              & 33.24                    &  28.75         &   21.16             \\
\multicolumn{1}{l|}{CASENN\cite{xue2021clustering}}                  & \textbf{57.46} & \textbf{49.44}         & 
\textbf{44.94}          & \multicolumn{1}{l|}{\textbf{40.71}}          & \textbf{56.96}          & \textbf{47.99 }         & \textbf{40.54}          & \textbf{36.54}          \\
\multicolumn{1}{l|}{COMPLETER\cite{lin2021completer}}              & 34.47          & 31.79          & 30.35          & \multicolumn{1}{l|}{31.30}           & 26.48          & 20.93          & 17.54          & 17.17          \\
\multicolumn{1}{l|}{CIGIT-C}                & 54.71          & \textbf{50.52} & \textbf{46.68} & \multicolumn{1}{l|}{\textbf{42.22}} & \textbf{56.83} & \textbf{42.84} & \textbf{42.11} & \textbf{39.92} \\ \hline
\end{tabular}}
\end{table*}

\begin{table*}[h]
\centering
\caption{Clustering results (ACC and NMI) of different methods on the NUS-WIDE-10K dataset with different missing rate ${p\%}$.}
\label{7}
\setlength{\tabcolsep}{6.0mm}{
\begin{tabular}{lllllllll}
\hline
\multicolumn{1}{c}{\multirow{2}{*}{Method}} & \multicolumn{4}{c}{Accuracy}                                                           & \multicolumn{4}{c}{NMI}                                           \\ 
\multicolumn{1}{c}{}                        & 10${\%}$               & 30${\%}$             & 50${\%}$                & 70${\%}$                                     & 10${\%}$               & 30${\%}$                & 50${\%}$                & 70${\%}$              \\ \hline
\multicolumn{1}{l|}{BestSM}                 &50.82         &47.14         &42.49        &\multicolumn{1}{l|}{32.88}          &57.71           & 48.76        &40.41            & 33.25         \\
\multicolumn{1}{l|}{DAIMC\cite{ijcai2018-313}}                  &61.84           &51.06            &39.04           & \multicolumn{1}{l|}{26.49}          & 51.75         &35.25        &24.21          & 17.12         \\
\multicolumn{1}{l|}{UEAF\cite{wen2019unified}}                   & 60.47         & 51.03         & 40.91         & \multicolumn{1}{l|}{32.18}          &56.72           & 45.54         & 34.12           & 25.42        \\
\multicolumn{1}{l|}{OPIMCC\cite{hu2019one}}                  &51.90         &35.91         &23.67         & \multicolumn{1}{l|}{15.95}           &55.60        & 45.72        & 40.76         &33.34          \\
\multicolumn{1}{l|}{PIC\cite{wang2019spectral}}                    & 51.52         & 47.09         & 36.73       & \multicolumn{1}{l|}{32.91}          &40.41         & 34.91       &  20.94        & 17.47        \\
\multicolumn{1}{l|}{CDIMC-net\cite{ijcai2020-447}}              & 59.98         & \textbf{58.41}           &  \textbf{51.28}       & \multicolumn{1}{l|}{\textbf{47.94}}         & 54.46               & \textbf{52.82}                &  \textbf{49.68}               & \textbf{34.23}                \\
\multicolumn{1}{l|}{CASENN\cite{xue2021clustering}}                  & \textbf{62.39}     & 56.34          & 50.74        & \multicolumn{1}{l|}{46.51}          & \textbf{62.63}        & 52.81          &  47.97       & 31.24         \\
\multicolumn{1}{l|}{COMPLETER\cite{lin2021completer}}              &  48.66        & 43.79        &  42.88        & \multicolumn{1}{l|}{39.67}           &  40.66      & 36.66        &39.51        & 33.17          \\
\multicolumn{1}{l|}{CIGIT-C}                &  \textbf{63.26}          & \textbf{57.44}  & \textbf{52.48} & \multicolumn{1}{l|}{\textbf{48.39}} & \textbf{63.56} & \textbf{54.93} & \textbf{49.65} & \textbf{38.73} \\ \hline
\end{tabular}}
\end{table*}

\

The real number of clusters  ${k}$  is set as the true number of classes for all data sets under the assumption that it is known.

We initialise 20 times and choose the best answer for each approach using the k-means algorithm to produce clustering assignments. In order to prevent extreme occurrences, we run each method 10 times and present the average results.

For DAIMC, we set
the parameters of the method by following the setting of
the original paper.

For UEAF, we set parameters ${\gamma}$ to 3. The role of parameter ${\gamma}$ is to control the smoothness of the weight distribution.

For OPIMC, in hyperparameter search, we select the parameter ${\alpha}$ within ${\{1 e^{-4},1 e^{-3},1 e^{-2},1 e^{-1},1 e^{0},1 e^{1},1 e ^{2},1 e ^{3}\}}$ and report the best results.

For PIC, we set parameters ${\tilde{\beta}}$ to 0.1.

For CDIMC-net and  CASEN, we choose the recommended network structure according to the original paper.

For CDIMC-net, the graph embedding
hyper-parameter ${\alpha}$ is fixed to ${1e^{-5}}$, and the learning rate   is fixed to ${1e^{-3}}$.

For CASEN, the important parameters are set as ${\eta_{1}=0.1}$, ${\eta_{2}=0.01}$, ${\lambda=0.5}$, ${\theta=0.1}$, ${r=2}$.

For COMPLETER, the softmax activation function is used at the
last layer of the encoders. The entropy parameter ${\alpha}$ is fixed to 9 and trade-off
hyper-parameters ${\lambda_{1}}$ and ${\lambda_{2}}$ are fixed to 0.1.

For CIGIT-C, the proposed method contains two training modules, 
i.e., modality-specific encoders and subspace conditional clustering GAN. Tables \ref{Table3}, \ref{Table4} and \ref{Table5} present the details of the network architectures in these two training modules, respectively. Take the Wikipedia dataset, ${\mathbf{E}^{(img)}}$ and ${\mathbf{E}^{(txt)}}$ are three-layer networks in our implementation. 
We set the dimension of the ${\mathbf{E}^{(img)}}$ to 4096${-}$1024${-}$256,  where 4096 is the feature dimension of the visual modality, 1024 denotes the hidden layer dimension,
256 denotes the subspace representation dimension. We set the dimension of the ${\mathbf{E}^{(txt)}}$ to 728${-}$256${-}$128. ${G_{12}}$  and ${G_{21}}$ are three-layer neural networks with a batch normalization layer to normalize the input vector and stabilize the training procedure. The proposed ${D_{1}}$  and ${D_{2}}$  are mainly made up of a fully connecteded layer with ReLU \cite{nair2010rectified} activation, a mini-batch layer that can increase the diversity of fake representations, a sigmoid function which outputs the fake-real possibility of input representations.
\begin{table*}[h]
\centering
\caption{Clustering results (ACC and NMI) of different methods on the BDGP dataset with different missing rate ${p\%}$.}
\label{8}
\setlength{\tabcolsep}{6.0mm}{
\begin{tabular}{lllllllll}
\hline
\multicolumn{1}{c}{\multirow{2}{*}{Method}} & \multicolumn{4}{c}{Accuracy}                                                           & \multicolumn{4}{c}{NMI}                                           \\ 
\multicolumn{1}{c}{}                        & 10${\%}$               & 30${\%}$             & 50${\%}$                & 70${\%}$                                     & 10${\%}$               & 30${\%}$                & 50${\%}$                & 70${\%}$              \\ \hline
\multicolumn{1}{l|}{BestSM}                 & 47.52      &  41.44      &  34.47       & \multicolumn{1}{l|}{27.95}          & 35.74        & 25.20       &16.39           & 9.29       \\
\multicolumn{1}{l|}{DAIMC\cite{ijcai2018-313}}                  & 74.76         & 62.88          & 52.45        & \multicolumn{1}{l|}{35.82}          &  55.64       & 47.87     & 28.33         & 9.17        \\
\multicolumn{1}{l|}{UEAF\cite{wen2019unified}}                   & 76.56       &  57.13         & 50.41        & \multicolumn{1}{l|}{36.84}          & 65.13        & 44.10      & 33.13        &  12.60      \\
\multicolumn{1}{l|}{OPIMCC\cite{hu2019one}}                  & 71.69         &  56.44       &  55.17       & \multicolumn{1}{l|}{35.82}           &61.77       & 41.47      & 25.94       &8.35      \\
\multicolumn{1}{l|}{PIC\cite{wang2019spectral}}                    & 68.26        &63.05        & 53.23       & \multicolumn{1}{l|}{44.31}          & 66.70        &  \textbf{55.77}   &  27.78       &  11.43     \\
\multicolumn{1}{l|}{CDIMC-net\cite{ijcai2020-447}}              &  \textbf{87.50}       &  \textbf{76.99}     &  \textbf{60.47}       & \multicolumn{1}{l|}{40.98}          &   \textbf{77.24}             & \textbf{57.09}           & \textbf{35.64}               &  \textbf{14.77}             \\
\multicolumn{1}{l|}{CASENN\cite{xue2021clustering}}                  & \textbf{79.92}  & 70.96           & 59.25    & \multicolumn{1}{l|}{46.93}          & \textbf{72.21}     &  48.23      & 29.77        &   12.45      \\
\multicolumn{1}{l|}{COMPLETER\cite{lin2021completer}}              &  59.60      & 55.20      &  54.10       & \multicolumn{1}{l|}{\textbf{52.90}}          & 52.80    &  51.10     & 33.04     & \textbf{14.77}          \\
\multicolumn{1}{l|}{CIGIT-C}                &78.12          & \textbf{72.14}  & \textbf{62.74} & \multicolumn{1}{l|}{\textbf{51.82}} & 68.41 & 51.76 & \textbf{36.11} & \textbf{16.78} \\ \hline
\end{tabular}}
\end{table*}

\subsection{Experimental Results}
Tables \ref{6} to \ref{8} list the clustering performance of all the compared methods on the three datasets, from which we can observe the following:

1) It is clear that when the missing rate changes from 0.1 to 0.7, the clustering performance of all approaches is decreased.

2) Although CIGIT-C  performed moderately compared to other baselines with a low miss rate, it was not outstanding. Compared with the second-best methods,  CIGIT-C has considerable improvements on Wikipedia, NUS-WIDE-10K, and BDGP dataset, when the missing rate is 50\% or 70\%. Specifically, for Wikipedia datasets, CIGIT-C achieves a 0.78\%, 1.74\%, and 1.64\% increment at a  miss rate of 30\%, 50\%, and 70\%, with respect to accuracy against the second best. For the indicator NMI, CGIG-C is higher than 1.57\% and 3.38\% compared to the second best baseline at a miss rate of 50\% and 70\%. For missing rate ${p\%=70}$, compared to the second best method,  CIGIT-C improves 
NMI by 4.5\%, 2.01\% on NUS-WIDE-10K  and BDGP datasets, respectively, which demonstrates the effectiveness of  CIGIT-C.

3) For Wikipedia dataset, some multi-view clustering methods (i.e., CDIMC-net and COMPLETER 
)  results even perform worse than single-view methods (i.e., BestSM) results.

4)  For BDGP with low missing rates, the performance of
certain methods (like OPIMC and COMPLETER) is bad.

The reasons for the above observations can be explained
as follows:

1) As the missing rate rises, complementary information between multi-modalities becomes scarce, and all methods inevitably decline.

2) CIGIT-C  methods are superior in most databases, especially
when the partial ratio is large. The reasons are 2.1) our model is generative and serves to augment the data, and the augmented data is helpful for the clustering task. 2.2) the model exploits potential representations within and between modalities. 2.3) our model not only generates data but also introduces a KL loss function to achieve predictive consistency across heterogeneous modalities.

3) For Wikipedia dataset, the performance of
certain methods (like CDIMC-net  and COMPLETER) is bad. The possible reasons are as follows: 3.1) Both method belong to the imcomplete multi-view method and do not consider the heterogeneity between images and text. 3.2) COMPLETER aims to maximise the mutual information between different views through contrastive learning to obtain a common representation to fill in the missing data. In the case of high missing rates, it is difficult to find connections between modalities.
3.3) The CDIMC-net padding strategy depends on the estimation of the data distribution. As a result, the cumulative error increases when the missing rate is high.

4) The similarity of the data is the primary variable influencing clustering performance. The critical component is still the multi-modal learning of features since similarity depends on characteristics. The low quality of the modal features in the BGDP dataset has a direct impact on the performance of the method.

5) DAIMC and OPIMC are based on matrix factorization. However, their clustering performance is limited because they mainly make use of some regularization and add some constraints to the new representation but fail to compensate for the missing data in each view explicitly.

6) UEAF and PIC explore intra-modality connections and structure for multi-view clustering yet neglect to explore the underlying relationships and structure between modalities. Therefore valid clustering results can not be obtained.

7) The direct and brutal fusion of heterogeneous data inevitably leads to performance degradation. The introduction of KL-divergence loss in  CIGIT-C largely circumvents the inconsistency of predictions across modal clusters by means of consistency constraints.

Figures \ref{Fig5} and \ref{Fig6} show the ACC and NMI metric values for all comparison methods on the three datasets at missing rates from 0.1 to 0.9 with an interval of 0.1, respectively.

Through these results, we can draw the following
conclusions:  
1) CIGIT-C methods are superior in most databases, especially when the missing ratio is large, i.e. missing rates ${p\%}$ from 50 to 90. For example, as shown in figure \ref{Fig5}(b), at a missing rate of 90, CGIT-C surpasses the optimal baseline by 5.57\%.
2)  With increasing the missing rate, the performance degradations of
the compared methods are much larger than that of ours. For example, when the missing rate is 50\%, as shown in Figure \ref{Fig5}(a) , CASEN and CGIG-C reach 44.94\% and  46.68\%, respectively. However, as the missing rate rise, CGIG-C is superior to CASEN.

In addition, as shown in Figure  \ref{Fig5}(a), an interesting phenomenon is that the results of some incomplete multi-view (UEAF, COMPLETER,DAIMC,CDIMC-net) methods are even worse than those of single-view(BestSM) methods, which also proves from the experimental results that the existing incomplete multi-view methods cannot effectively solve IITC tasks.

\begin{figure*}[t]
\label{Fig5}
  \begin{center}
  \includegraphics[width=0.90\textwidth]{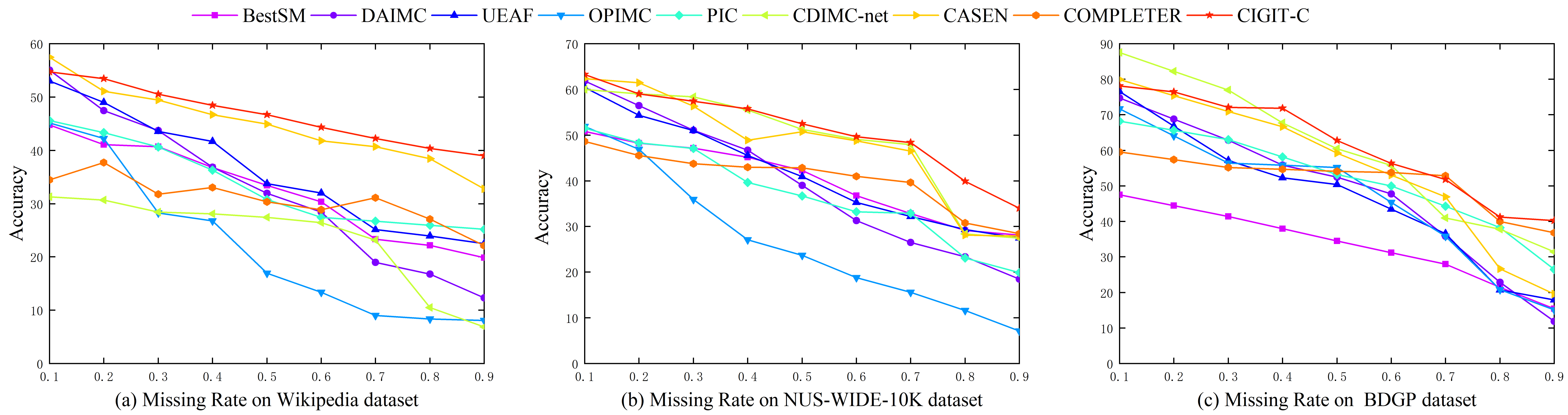}
  \caption{The average clustering accuracy performance with respect to different missing rate on the (a) Wikipedia dataset, (b)  NUS-WIDE-10K dataset  and (c) BDGP dataset.}\label{Fig5}
  \end{center}
\end{figure*}

\subsection{Parameter Analysis and Ablation Study}
This section examines the CIGIT-C on the Wikipedia, NUS-WIDE-10K, and BDGP datasets from the perspective of parameter analysis and ablation studies.
Our method contains four user-defined parameters, weighting factor ${\beta}$ to balance the ratio of image modality to text modality and weighting factor ${\alpha}$ to control KL loss. ${\eta _{m}}$ is the generated and true representation weighting coefficient for the m-th order modality.  ${\eta _{1}}$  is for image modality, and  ${\eta _{2}}$ is for text modality.

As shown in Figure 7(a), when the missing rate is 0.5, parameter ${\beta}$ has the best effect when it is 0.6, 0.4 and 0.4 on the three datasets of Wikipedia, NUS‐WIDE‐10K and BDGP, respectively. The reason for the different values of parameter ${\beta}$ may be that, in contrast to the Wikipedia dataset, the text features extracted from the tags are very simple. Therefore, the semantic information provided by text modalities in the other two datasets is weaker than that provided by text modalities in the Wikipedia dataset. The parameter ${\alpha}$ is tuned from $\{{0.01,0.1,1,10,100\}}$ and the ACC performance is shown in Figure 7(b).  

The  ${\eta _{1}}$  and  ${\eta _{2}}$  parameters are tuned in the same range as ${\alpha}$. The best results are achieved on the Wikipedia dataset for a missing rate of 0.5, with both taking the value of 0.1. The parametric analysis of CIGIT-C on the NUS-WIDE-10K and BDGP datasets is analysed using the same methods. Due to space constraints, we will not continue our conversation and show here.

To demonstrate the importance of the Clustering GAN module and KL loss in CIGIT-C, we performed ablation experiments with a miss rate of 0.5. As shown in Figure \ref{Fig8}, where"None Clustering GAN" means that the conditional clustering GAN does not participate in the CIGIT-C work, similarly, "None KL-Divergence  losses" means that CGIG-C contains only the clustered GAN.

We can draw the following conclusions:

1) It can be seen that when both clustering GAN and KL-loss exist, CIGIT-C performs the best, indicating that modules and loss functions in CIGIT-C complement each other and promote each other.

2) The CIGIT-C model does not improve much when only Kl-loss is available. Even with the consistency constraint, i.e. the different and fused modalities predict the clustering knowledge consistently. However, the performance of CIGIT-C inevitably decreases because no treatment is given to the missing values.

3) When only clustering GAN is available, although clustering GAN serves to increase the diversity of cross-modal representations and solve the problem of missing modalities, it lacks consideration of the inconsistencies arising from inter-modal clustering predictions.
\begin{figure*}[t]
\label{Fig6}
  \begin{center}
  \includegraphics[width=0.90\textwidth]{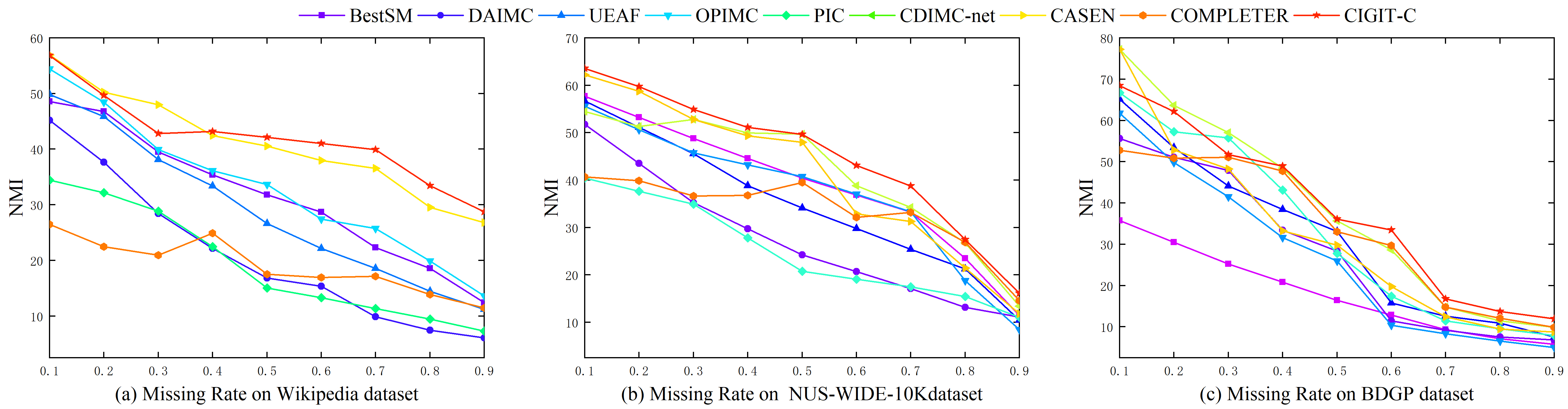}
  \caption{The average clustering NMI performance with respect to different missing rate on the (a) Wikipedia dataset, (b)  NUS-WIDE-10K dataset  and (c) BDGP dataset .}\label{Fig6}
  \end{center}
\end{figure*}

\begin{figure}[h]
\centering
 
  \includegraphics[width=0.95\linewidth]{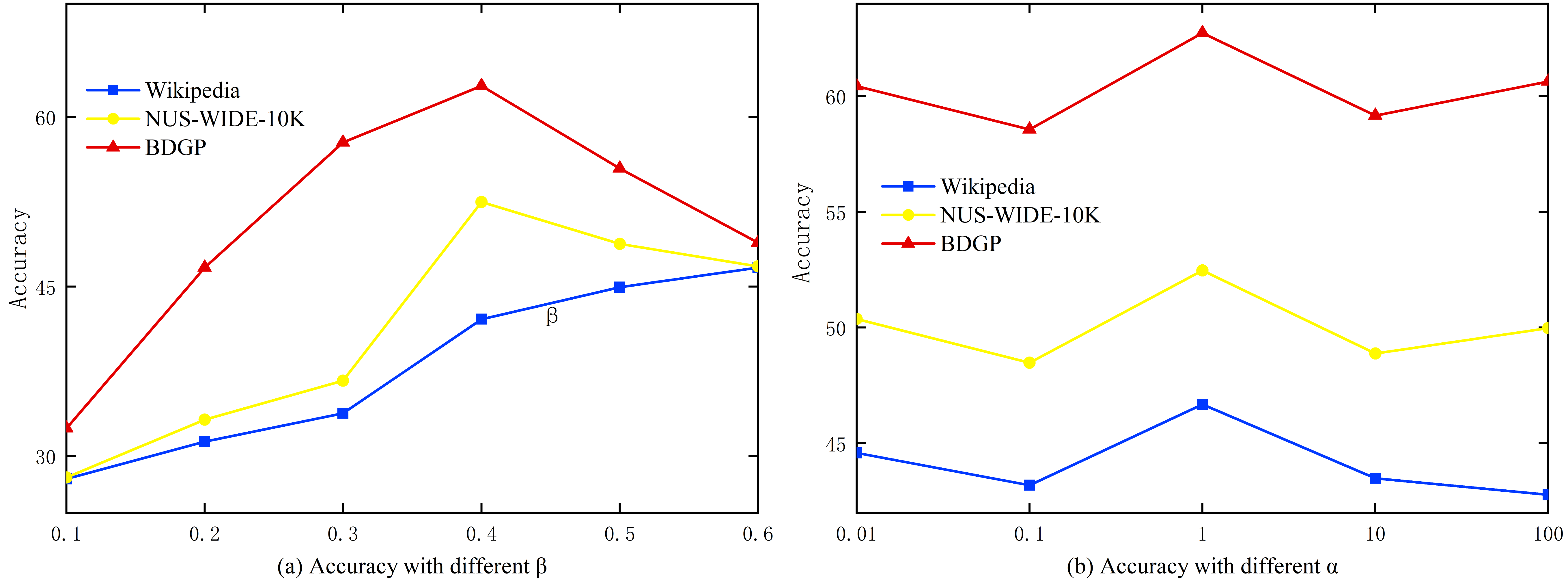}
  \caption{Accuarcy (\%) performance with different  ${\beta}$ and ${\alpha}$ when
the missing rate ${p}$ is 0.5.}
\label{Fig7}
\end{figure}

\begin{figure}[h]
\centering
 
 \includegraphics[width=0.95\linewidth]{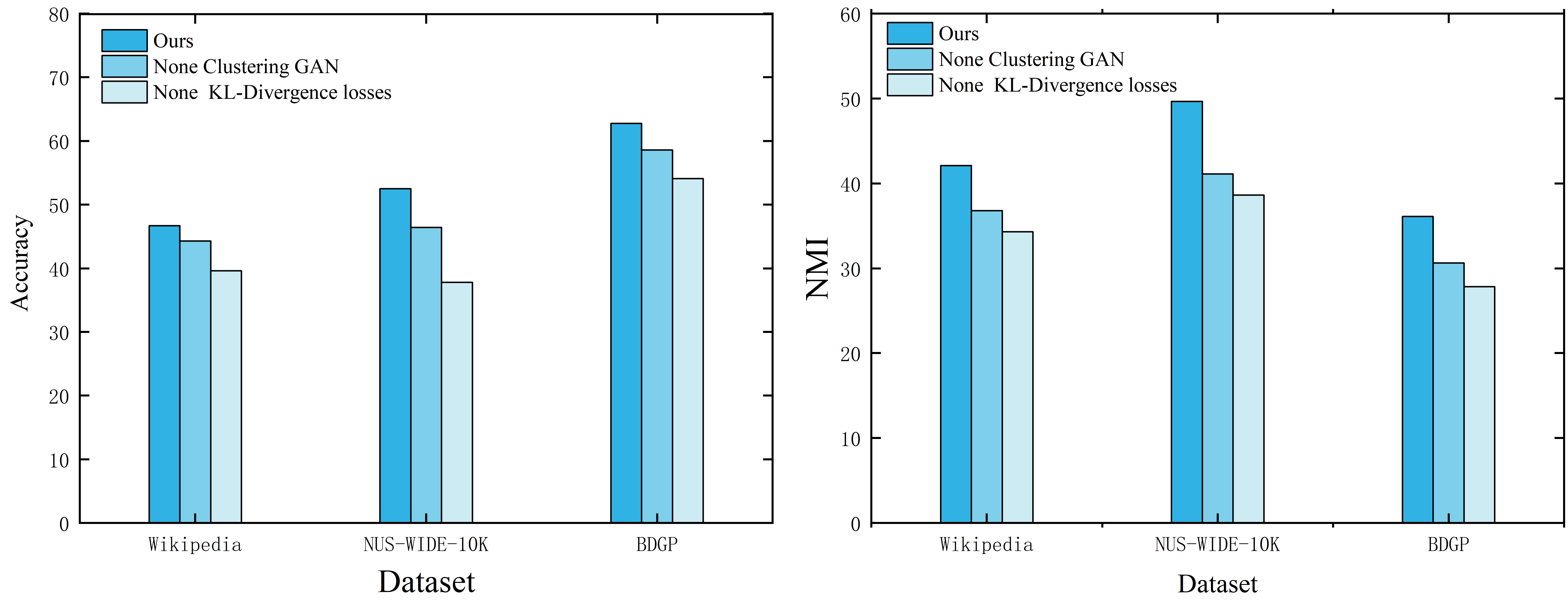}
  \caption{ Effectiveness of clustering GAN, KL-divergence losses, when the missing rate
${p}$ is 0.5.}
\label{Fig8}

\end{figure}

\begin{figure}[h]
\centering
\subfigure{
\includegraphics[width=6cm]{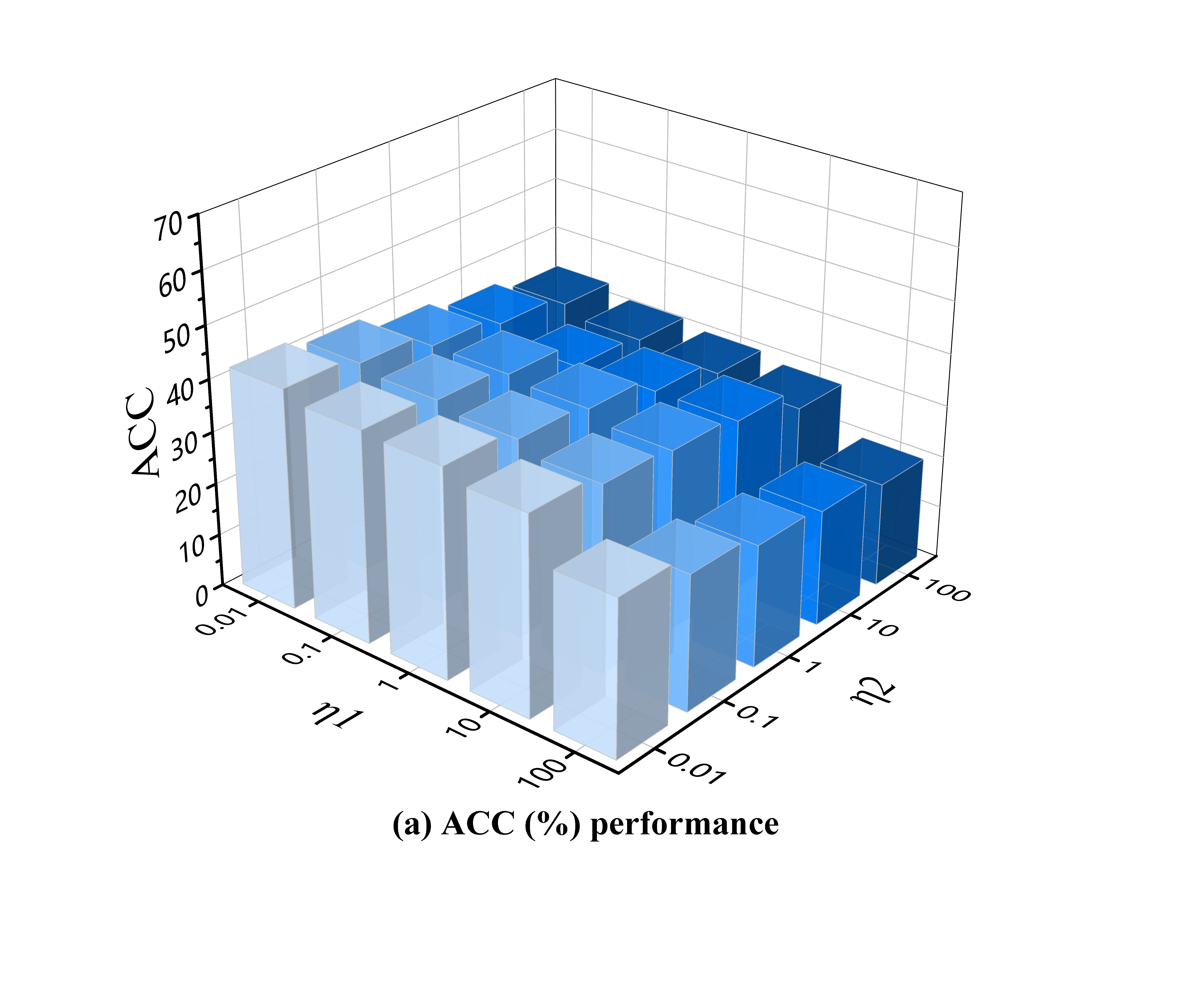}
%\caption{fig1}
}
\quad
\subfigure{
\includegraphics[width=6cm]{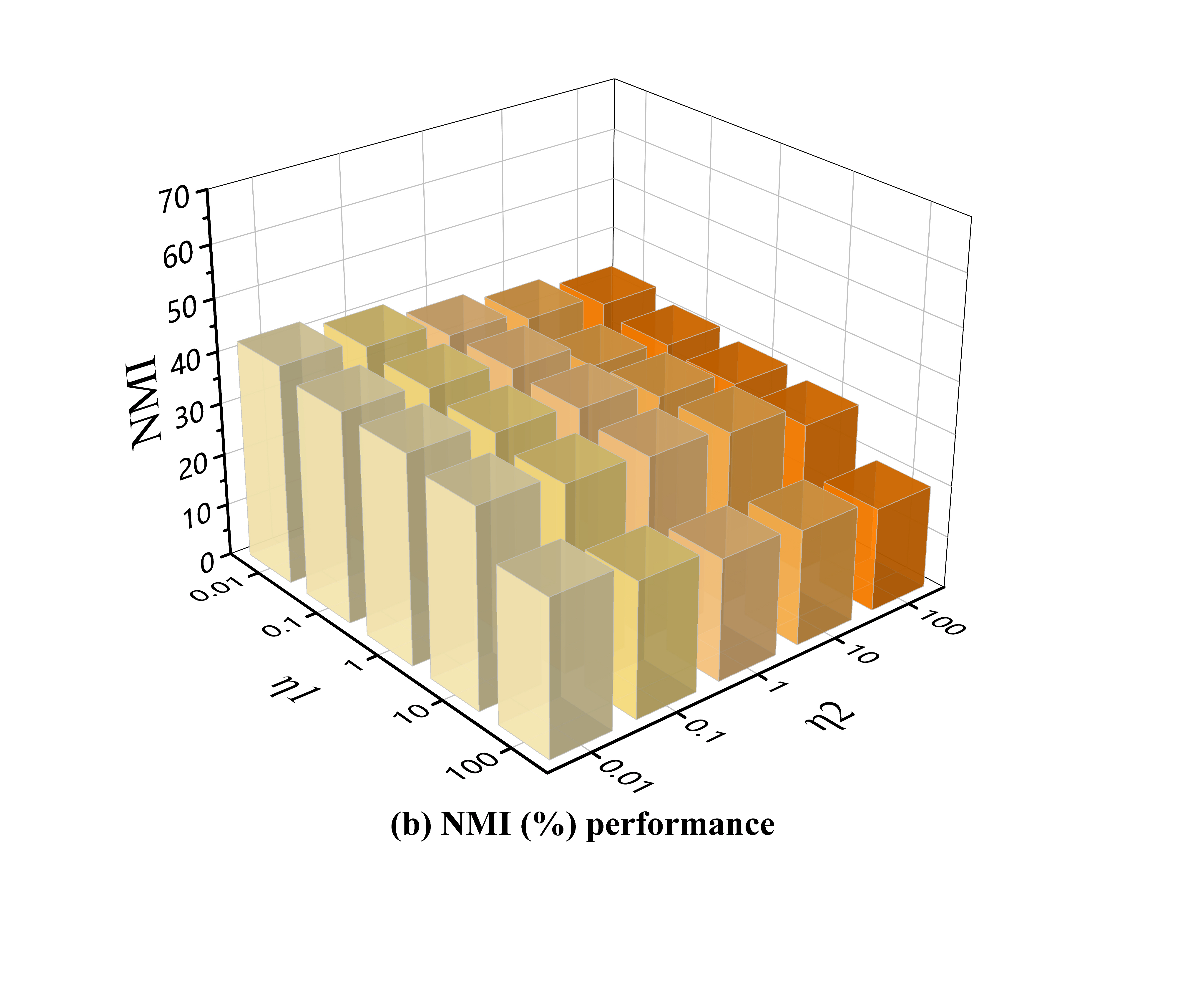}
}
\caption{Parameter analysis on Wikipedia dataset.}
\label{Fig10}
\end{figure}

\subsection{Qualitative Analysis}
\begin{figure}[h]
\centering
\subfigure[Real and Fake Samples in ${\mathbf{Z}^{(img)}}$]{
\includegraphics[width=3.1cm]{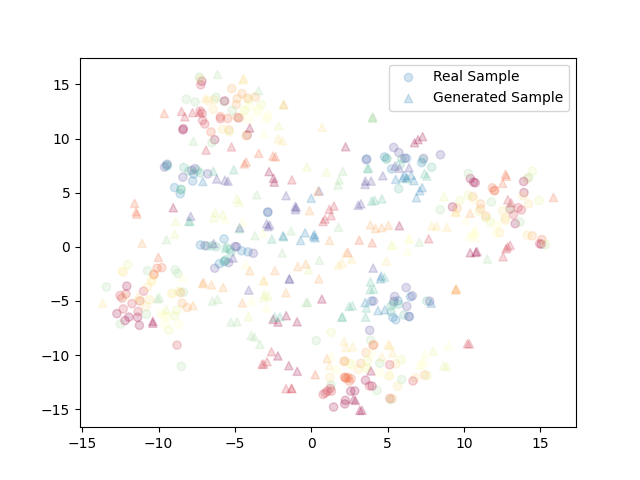}

}
\quad
\subfigure[Real and Fake Samples in ${\mathbf{Z}^{(txt)}}$]{
\includegraphics[width=3.1cm]{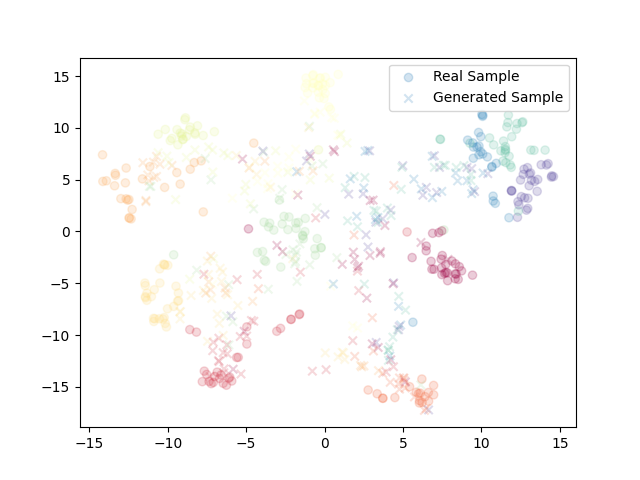}

}
\caption{ t-SNE \cite{van2008visualizing} visualization results of the real and the generated test sample representations in ${\mathbf{Z}^{(img)}}$  and ${\mathbf{Z}^{(txt)}}$  respectively.}
\label{Fig11}
\end{figure}

\begin{figure}[h]
\centering
\includegraphics[width=0.5\textwidth]{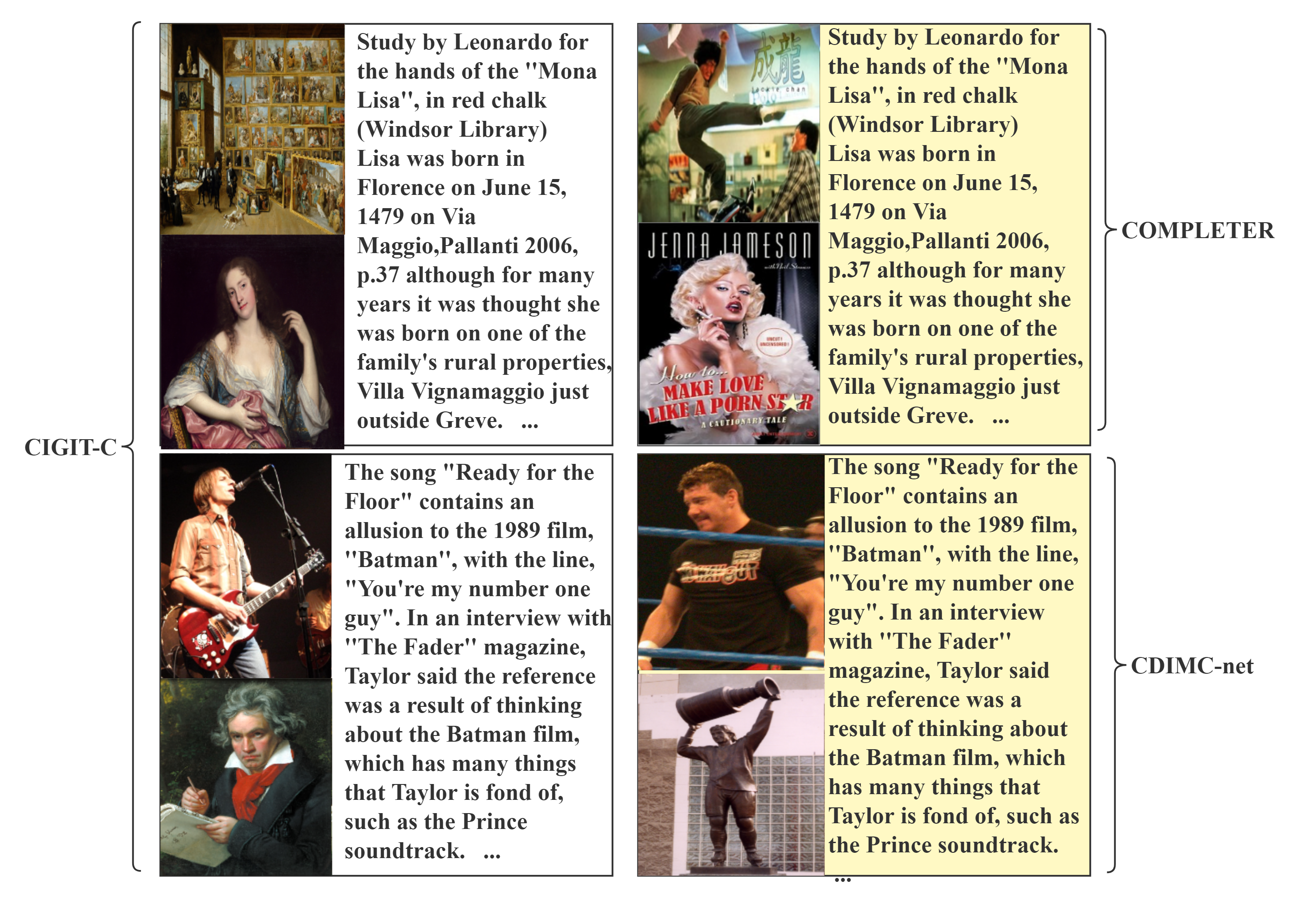}
\caption{The SOTA method and the CIGIG-C method visualise the clustering results. The bottom white clustering result in the figure is correct, with both images and text belonging to the same semantic category. The categories are art and music from top to bottom, while the yellow box shows the wrong clustering results.}
\label{Fig12}
\end{figure}

In order to verify the effectiveness of clustering GAN in CIGIT-C, we visualize the distribution of real and generated representations of samples in ${\mathbf{Z}^{(img)}}$ and ${\mathbf{Z}^{(txt)}}$  by t-SNE \cite{van2008visualizing}. The results are shown in Figure \ref{Fig11}, which illustrates that the true and generated representations belonging to the same cluster category are close to each other and vice versa. The visualisation in Figure \ref{Fig12} demonstrates the advantages of our proposed CIGIT-C in terms of clustering results.
In contrast to CGIG-C, COMPLETER\cite{lin2021completer} incorrectly clusters text with the semantic category art and media images together. CDIMC-net\cite{ijcai2020-447} incorrectly clusters text with the semantic category music and sports images together.
\section{Conclusion}

We demonstrate theoretically and experimentally that the existing incomplete multi-view clustering method can not solve the IITC task well, which is not considered in the existing works. We propose a clustering Gan to ensure that the populated data benefits the clustering task. We also add consistency constraints by introducing the KL divergence loss function to reduce the inconsistency caused by different modality predictions.

\section*{Acknowledgments}
This work was supported in part by the Major Program of the Inner Mongolia Natural Science Foundation under Grant 2021ZD0015, in part by the Natural Science Foundation
of China under Grant 52069018, in part by Natural Science Foundation of Inner Mongolia under Grant 2020MS03009, in part by Higher Education Scientific Research Program of Inner Mongolia under Grant NJZY22370, in part by Science and Technology Program of Inner Mongolia under Grant 2020GG0078.

\medskip
\bibliography{ref}

\end{document}